\documentclass{article} 
\usepackage{iclr2027_conference,times}

\usepackage{hyperref}
\usepackage{url}

\title{Jaxolotl: A Unified High-Performance Benchmark Suite for LTL-Based Multi-Task RL}

\iclrfinalcopy

\author{%
Mathias Jackermeier$^{*1}$ \qquad Jacques Cloete$^{*2}$ \qquad Alessandro Abate$^1$\\
$^1$Department of Computer Science, University of Oxford\\
$^2$Oxford Robotics Institute, University of Oxford\\
}

\usepackage{lipsum}
\usepackage{amsmath}
\usepackage{amssymb}
\usepackage{amsthm}
\usepackage{algpseudocode}
\usepackage{algorithm}
\usepackage{booktabs}
\usepackage{tabularx}
\usepackage{makecell}
\usepackage{array}
\usepackage{nicefrac}
\usepackage{microtype}
\usepackage{xcolor}
\usepackage{mathtools}
\usepackage{graphicx}
\usepackage{multirow}
\usepackage{enumitem}
\usepackage{tcolorbox}
\usepackage{inconsolata}
\usepackage{comment}
\usepackage{bm}
\usepackage{array}
\usepackage[noabbrev, capitalize]{cleveref}
\usepackage{subcaption}
\usepackage{xspace}
\usepackage[bb=dsserif]{mathalpha}
\usepackage{tikz}
\usepackage{lipsum}
\usepackage[export]{adjustbox}
\usetikzlibrary{automata,arrows.meta,calc,positioning,decorations.pathreplacing,backgrounds,matrix,arrows,patterns}
\usepackage{wrapfig}

\newcommand{\jaxolotl}{\textsc{Jaxolotl}}

\newcommand*{\given}{\,|\,}
\newcommand{\refp}[1]{\S\,\ref{#1}}

\newcommand{\always}{\mathsf{G}\,}
\newcommand{\event}{\mathsf{F}\,}

\newcommand{\nex}{\mathsf{X}\,}
\newcommand{\until}{\;\mathsf{U}\;}

\newcommand{\gror}{\;|\;}

\DeclareMathOperator*{\argmax}{arg\,max}

\DeclareMathOperator*{\E}{\mathbb E}

\DeclarePairedDelimiter\abs{\lvert}{\rvert}%
\DeclarePairedDelimiter\norm{\lVert}{\rVert}%
\makeatletter
\let\oldabs\abs
\def\abs{\@ifstar{\oldabs}{\oldabs*}}
\let\oldnorm\norm
\def\norm{\@ifstar{\oldnorm}{\oldnorm*}}
\makeatother

\newcommand{\tikzxmark}{%
    \tikz[scale=0.23] {
        \draw[line width=0.7,line cap=round] (0,0) to [bend left=6] (1,1);
        \draw[line width=0.7,line cap=round] (0.2,0.95) to [bend right=3] (0.8,0.05);
    }}
\newcommand{\tikzcmark}{%
    \tikz[scale=0.23] {
        \draw[line width=0.7,line cap=round] (0.25,0) to [bend left=10] (1,1);
        \draw[line width=0.8,line cap=round] (0,0.35) to [bend right=1] (0.23,0);
    }}

\usepackage{pifont}

\newtcolorbox{terminalcmd}{
    colback=gray!6,          
    colframe=gray!30,
    arc=2.5pt,
    boxrule=0.5pt,
    left=8pt, right=8pt,
    top=5pt, bottom=5pt,
    width=\linewidth,
    fontupper=\small\ttfamily
}

\newcommand{\linefill}{%
  \leavevmode\leaders\hrule height 2.4pt depth -2.0pt\hfill\kern0pt%
}
\newcommand{\spanval}[1]{\linefill\enspace #1\enspace\linefill}

\newcommand{\codeurl}{\url{https://github.com/mathiasj33/jaxolotl}}

\definecolor{myBlue}{HTML}{245C8A}
\hypersetup{
	colorlinks,
	linkcolor={myBlue},
	citecolor={myBlue},
	urlcolor={myBlue}
}
\creflabelformat{equation}{#2\textup{#1}#3}

\begin{document}

\maketitle

\begin{abstract}
   Training agents to follow arbitrary instructions is an important goal of multi-task reinforcement learning (RL).
   Linear temporal logic (LTL) provides a precise and structured formalism for specifying instructions to agents, and has been successfully adopted for training generalist multi-task policies.
   However, differences in implementations, task distributions, and evaluation protocols make existing methods difficult to compare, while high computational costs limit the scale and statistical reliability of experiments.
   We introduce \jaxolotl, a unified high-performance benchmark suite for multi-task LTL-RL to address these concerns.\hyperlink{fn:code}{\footnotemark[1]}
   \jaxolotl\ provides a modular, end-to-end JAX implementation of six representative algorithms and four environments, together with newly curated task suites and a standardised, statistically robust evaluation protocol.
   By precompiling symbolic task representations into static arrays, \jaxolotl\ enables fully JIT-compiled training and evaluation, achieving end-to-end speedups of up to $220\times$ and supporting controlled comparisons at substantially greater experimental scale.
   We use this framework to systematically evaluate existing approaches, revealing complementary strengths and limitations:
   general methods capable of non-myopic reasoning struggle as the number of propositions grows, while methods with stronger scaling rely on environment-specific assumptions and suffer from myopia.
\end{abstract}

{%
  \renewcommand{\thefootnote}{*}%
  \footnotetext{Equal contribution\;\;\hypertarget{fn:code}{\textsuperscript{1}}Code is available at: \codeurl}%
}
\setcounter{footnote}{1}

\section{Introduction}
Multi-task reinforcement learning (RL) is a promising approach towards training generalist agents that can complete diverse objectives.
A central question in this setting is how to specify tasks (or \textit{instructions}) to the agent.
While many works have explored natural language as an intuitive interface~(e.g.\ \citealp{hill2020Human,carta2023Grounding}), there has been increasing interest in \textit{formal} specification languages such as \textit{linear temporal logic} (LTL;~\citealp{pnueli1977temporal}).
LTL specifies temporally extended tasks over \textit{atomic propositions}, i.e.\ high-level Boolean features of the environment state, and is especially appealing due to its unambiguous semantics and compositional task structure.

Despite recent progress in multi-task LTL-conditioned RL~\citep{vaezipoor2021LTL2Action,qiu2023Instructing,liu2024Skill, jackermeier2025DeepLTL,guo2025One,jackermeier2026ZeroShot}, it remains challenging to understand how different methods actually compare:
existing evaluations rely on different environments, evaluation protocols, and low-level implementation details, making it difficult to draw general conclusions.
As a result, it is unclear whether reported differences arise from algorithmic choices or from experimental setup.
Systematic comparison is further hindered by the computational cost of experiments, which typically involve training RL agents for millions of steps in CPU-based environments, restricting evaluations to a few random seeds and limiting broad analyses.

We introduce \jaxolotl~(\textbf{JAX}-\textbf{O}ptimised \textbf{L}earning for \textbf{O}bjectives in \textbf{T}emporal \textbf{L}ogic) as a unified benchmark suite to address these shortcomings.
\jaxolotl\ provides a modular, end-to-end JAX \citep{jax2018github} implementation of six representative algorithms and four environments, spanning high-dimensional navigation, continuous control, and object interaction.
We precompile inherently symbolic task structures, such as finite automata and syntax trees, into static tensor representations to meet JAX's strict requirements on static array shapes and functional control flow.
This allows us to fully just-in-time (JIT) compile the entire training and evaluation loop, leading to end-to-end speedups of up to $220\times$ over reference implementations on hardware accelerators.
These efficiency improvements enable controlled benchmarking at substantially greater experimental scale, such as training 10 independent seeds in under four minutes rather than 13.4 hours.

We use \jaxolotl\ to conduct an extensive controlled benchmark of existing methods, isolating aspects such as general task satisfaction, non-myopic reasoning, and scaling with the number of atomic propositions.
Our experiments reveal complementary limitations in current approaches: general methods capable of non-myopic reasoning (i.e.\ reasoning about the entire task rather than individual subgoals) struggle to scale to environments with large numbers of propositions.
Conversely, methods exhibiting stronger proposition scaling rely on environment-specific observation-reduction functions and suffer from myopia, preventing them from planning multiple steps ahead.

Our main contributions are as follows:
\begin{itemize}[itemsep=0em]
   \item we introduce \jaxolotl, an open-source JAX benchmark suite for multi-task LTL-RL, unifying six representative algorithms and four environments under shared abstractions;
   \item we design a precompilation strategy for dynamic symbolic task representations that enables end-to-end JIT compilation, yielding training speedups of up to $220\times$;
   \item we establish a standardised evaluation protocol and curated task suites for controlled comparison across methods, including uncertainty estimates over independently trained policies;
   \item lastly, we conduct an extensive benchmark evaluating existing methods across general task completion, non-myopic reasoning, and proposition scaling, uncovering complementary strengths and limitations of current approaches.
\end{itemize}

\section{Related Work}
\label{sec:related_work}

\begin{figure}
   \centering
   \includegraphics[width=0.92\textwidth]{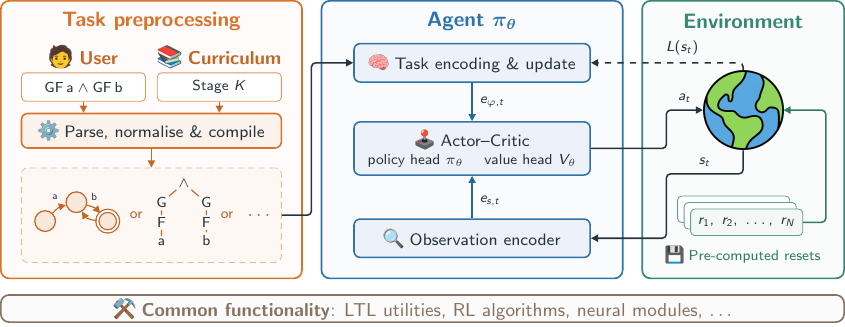}
   \caption{Overview of \jaxolotl's design and implementation. We provide a modular implementation of representative multi-task LTL-RL algorithms in terms of a generic task representation, policy architecture, and shared common modules. Precompilation of dynamic structures and environment resets enables end-to-end JIT compilation of the entire training and evaluation loop.}
   \label{fig:design}
\end{figure}

\paragraph{LTL-Based Multi-Task RL.}
LTL and related formalisms such as reward machines are widely used to specify single, fixed tasks for RL agents \citep{hahn2019OmegaRegular,camacho2019LTL,toroicarte2022Reward,voloshin2023Eventual,hasanbeig2023Certified}.
We instead consider the \textit{multi-task} setting, where a single policy is required to zero-shot execute arbitrary LTL instructions.
Many approaches have been developed for this setting, which can be broadly categorised into \textit{decomposition-based} and \textit{holistic} methods.
Decomposition-based methods break down LTL specifications into subgoals that can be completed one at a time by a goal-conditioned policy~\citep{araki2021Logical,leon2022Nutshell,qiu2023Instructing,liu2024Skill,guo2025One}.
While this simplifies learning, it limits the agent's ability to reason about the entire task and may lead to myopic behaviour.
Holistic approaches instead condition the policy on broader task structure, for example by encoding the formula syntax tree~\citep{vaezipoor2021LTL2Action} or B\"uchi automaton sequences~\citep{jackermeier2025DeepLTL}.
Recent work targets scaling to large proposition sets, via proposition-specific observation reduction \citep{guo2025One} or parameterised predicates \citep{cloete2026PlatoLTL}.
While the field has seen rapid progress in recent years, it remains difficult to compare methods due to differences in environments, task distributions, and evaluation protocols.
Our work builds on these existing LTL-based multi-task RL methods and unifies them within a single JAX framework.

\paragraph{Existing LTL-Based Benchmarks.}
Closest to our work, \citet{guo2026SpecRLBench} introduce SpecRLBench, a suite of environments and tasks for generalisation in multi-task LTL-RL.
While SpecRLBench standardises several experimental settings, it relies on the methods' original codebases, retaining differences in their training and evaluation pipelines.
In contrast, \jaxolotl\ provides a unified implementation of six representative algorithms, together with a standardised evaluation protocol and curated task suites for controlled comparison.
Furthermore, while SpecRLBench relies on CPU-based environments, \jaxolotl\ enables end-to-end training and evaluation on hardware accelerators, yielding orders-of-magnitude speedups over the reference algorithm implementations.
These speedups make larger-scale, statistically robust benchmarking practical.
Concurrent work~\citep{peng2026ManiGuard} evaluates VLA policies using LTL$_f$ monitors, but does not address multi-task LTL-RL methods.

\paragraph{Hardware-Accelerated RL.}
The speedups enabled by end-to-end JIT-compiled training in JAX~\citep{lu2022Discovered} have spurred a broad ecosystem of accelerated environments \citep{freeman2021Brax,lange2022Gymnax,bonnet2024Jumanji,matthews2024Craftax,pignatelli2025NAVIX,zakka2025MuJoCo} and algorithm libraries \citep{liesen2024Rejax,rutherford2024JaxMARL}.
Closest to our design, XLand-MiniGrid \citep{nikulin2024XLandMiniGrid} encodes procedurally generated symbolic rules and goals as fixed-size arrays to enable compiled multi-task training.
However, these accelerated frameworks do not support LTL-specified tasks, whose inherently dynamic structure in the form of finite automata or syntax trees poses a significant challenge for JIT compilation.
See \cref{app:extended_related_work} for extended related work.

\section{Background}

\paragraph{Reinforcement Learning.}
We consider a standard reinforcement learning (RL) setting, where an agent interacts with an environment modelled as a Markov decision process (MDP).
An MDP is a tuple $\mathcal{M} = (\mathcal{S}, \mathcal{A}, P, R, \gamma, \rho_0)$, where $\mathcal{S}$ is the state space, $\mathcal{A}$ is the action space, $P: \mathcal{S} \times \mathcal{A} \times \mathcal{S} \rightarrow [0, 1]$ is the transition kernel, $R: \mathcal{S} \times \mathcal{A} \times \mathcal S \rightarrow \mathbb{R}$ is the reward function, $\gamma \in [0, 1)$ is the discount factor, and $\rho_0 \in \Delta(\mathcal{S})$ is the initial state distribution. The agent's objective is to learn a (stationary) policy $\pi: \mathcal{S} \rightarrow \Delta(\mathcal{A})$ that maximises the expected cumulative discounted reward $\mathbb{E}_{\tau \sim \pi} \left[ \sum_{t=0}^{\infty} \gamma^t R(s_t, a_t, s_{t+1}) \right],$
where $\tau \in (\mathcal S\times\mathcal A)^\infty$ is a trajectory generated by $\pi$ in $\mathcal{M}$.

\paragraph{Linear Temporal Logic.}
Linear temporal logic (LTL;~\citealp{pnueli1977temporal}) is a formal language for specifying temporal properties of systems, which is being increasingly adopted as a task specification language in multi-task RL\@.
The basic unit of LTL formulae are atomic propositions $AP$, which represent Boolean properties of the environment (e.g.\ ``the agent is at the goal'').
The syntax of LTL formulae is defined recursively as follows:
$$ \varphi ::= \top \gror \mathsf p \gror \neg \varphi \gror \varphi_1 \land \varphi_2 \gror \nex \varphi \gror \varphi_1 \until \varphi_2 $$
where $\top$ is the Boolean constant $\mathsf{true}$, $\mathsf p \in AP$ is an atomic proposition, $\neg$ is negation, $\land$ is conjunction, $\nex$ is the next operator, and $\until$ is the until operator.
Standard Boolean connectives (e.g.\ $\lor$, $\rightarrow$) and temporal operators (e.g.\ $\event$, $\always$) can be derived from the above syntax.
Intuitively, $\nex \varphi$ means that $\varphi$ is true in the next step, $\varphi_1 \until \varphi_2$ means that $\varphi_1$ holds until $\varphi_2$ is true, $\event \varphi$ means that $\varphi$ eventually holds at some future time step, and $\always \varphi$ means that $\varphi$ is true always (at all time steps).
These operators allow for the specification of complex compositional and temporally extended tasks, such as ``eventually reach the goal while avoiding obstacles'' or ``visit all rooms in a specific order''.

To interpret LTL formulae in an MDP, we assume a labelling function $L\colon \mathcal{S} \rightarrow 2^{AP}$ that returns the set of true atomic propositions for a given state.
An MDP trajectory $\tau$ satisfies an LTL formula $\varphi$, denoted by $\tau \models \varphi$, if the sequence of labels $L(s_0), L(s_1), \ldots$ satisfies the formula according to the standard semantics of LTL (see \cref{app:ltl-semantics})\@.
Multi-task RL with LTL specifications aims to train generalist LTL policies by solving the following optimisation problem:
\begin{equation*}
   \pi^*(\cdot\given\varphi) \in \argmax_\pi \E_{\substack{\varphi \sim \mathcal D, \\ \tau \sim \pi \given \varphi}}\big[ \mathbb{1}[ \tau \models \varphi ] \big],
\end{equation*}
where $\mathcal D$ is an arbitrary distribution over LTL tasks, and
$\pi\given\varphi$ is a policy conditioned on the task $\varphi$.


\section{Jaxolotl}
We present \textsc{Jaxolotl} (\textbf{JAX}-\textbf{O}ptimised \textbf{L}earning for \textbf{O}bjectives in \textbf{T}emporal \textbf{L}ogic), a unified high-performance benchmark suite for multi-task RL with LTL specifications.
\jaxolotl\ provides a modular end-to-end JAX implementation of representative multi-task LTL-RL environments and algorithms, enabling controlled benchmarking at substantially greater experimental scale.

\subsection{Design Principles}\label{sec:design}
See \cref{fig:design} for an illustration of \jaxolotl's design and implementation principles.
The framework is designed to be modular via common abstractions, leverages flexible configuration management, and achieves highly efficient training and evaluation via end-to-end JIT compilation.

\paragraph{Modular Abstractions.}
\jaxolotl\ defines modular abstractions that allow implementing LTL-RL methods in terms of a generic \textit{task representation} and \textit{policy architecture}.
Here, a task representation is the algorithm-specific data structure used to encode the current LTL instruction, such as a formula syntax tree~\citep{vaezipoor2021LTL2Action}, reach--avoid sequence~\citep{jackermeier2025DeepLTL}, or reach--avoid subgoal~\citep{guo2025One}.
As the agent interacts with the environment and completes parts of the task, this representation is generally updated online.
A \textit{task encoder} maps this representation to an embedding for the policy.
The policy architecture further consists of an \textit{observation encoder} to process environment observations, and a shared \textit{actor--critic} module that outputs actions and value estimates based on the combined observation and task embeddings.

Actions and value estimates are used for policy optimisation using goal-conditioned RL~\citep{liu2022GoalConditioned}, where goals are task representations sampled at the beginning of each episode.
A reusable \textit{curriculum learning} implementation lets users gradually increase task complexity during training.
Curricula consist of a sequence of stages with different task distributions, and the agent progresses to the next stage once it achieves a specified performance threshold.
\jaxolotl\ currently supports on-policy optimisation with PPO~\citep{schulman2017proximal}, but the abstractions are agnostic to the underlying RL algorithm and can be extended to support off-policy methods.

By providing reusable implementations of common modules, \jaxolotl\ enables controlled comparisons between different algorithms, while keeping shared components such as observation encoders fixed.
The modular design further makes it easy to combine components from different methods, and to extend the framework with new approaches in the future.

\paragraph{Flexible Configuration.}
Experiments are configured using Hydra~\citep{yadan2019Hydra}, a flexible configuration management system.
We provide pre-defined configurations for the entire benchmark suite, and individual hyperparameters or components can be overridden directly from the command line.
To illustrate, training DeepLTL~\citep{jackermeier2025DeepLTL} on ZoneEnv is as simple as running:
\begin{terminalcmd}
   \textcolor{gray}{\$ }python scripts/train.py alg=deep\_ltl env=zone\_env
\end{terminalcmd}

\paragraph{End-to-End JIT Compilation.}
JAX's just-in-time (JIT) compilation achieves significant speedups by compiling Python code into XLA-optimised kernels for accelerators such as GPUs or TPUs.
\jaxolotl\ follows the recent trend of JIT-compiling the \textit{entire} training and evaluation loop~\citep{lu2022Discovered,liesen2024Rejax}, rather than only the environment step function or policy network.

However, end-to-end compilation is considerably more challenging for LTL-RL than for standard RL workloads.
The algorithms considered in this work manipulate symbolic objects whose structure depends on the sampled task, including formula syntax trees, automata with variable topology, and reach--avoid sequences with variable numbers of steps.
These task representations are inherently dynamic and may change during environment interaction, directly conflicting with JAX's strict requirements for static tensor shapes and control flow during JIT compilation.\footnote{See \url{https://docs.jax.dev/en/latest/notebooks/Common_Gotchas_in_JAX.html} for an overview.}

\jaxolotl\ addresses this mismatch by extensively \textit{precompiling} dynamic structures into static array representations, such as padded adjacency lists for formula graphs or padded transition tables for B\"uchi automata.
Before training, we sample tasks for each curriculum stage, convert them into method-specific array representations, and pad all variable-sized components to fixed shapes.
Task sampling and updating then reduce to simple array indexing operations, which are fully compatible with JIT compilation.
We similarly precompute compact environment-reset descriptors to efficiently handle control flow under JIT, and compile evaluation formulae into static representations.
The entire training and evaluation loops are therefore pure JIT-compiled functions, which further enables straightforward parallelisation across independent seeds via \texttt{vmap}.

\subsection{Evaluation Infrastructure}
\label{sec:eval_infra}
Deep RL algorithms are well-known to be sensitive to hyperparameter choices and random initialisation~\citep{henderson2018Deep,chan2019Measuringa}, making meaningful comparisons between different methods challenging.
\jaxolotl's computational efficiency allows increasing the number of independent training runs and evaluation episodes at marginal cost (see \S\,\ref{sec:efficiency}), thereby improving estimates of expected performance and its uncertainty due to training randomness.
We leverage this to provide a standardised evaluation protocol that adapts the general evaluation principles of~\citet{agarwal2021Deep} to the generalist-policy setting.

In particular, \jaxolotl\ reports estimates of expected performance on a fixed
set of benchmark tasks, together with 95\% confidence intervals~(CIs) over
independent training runs.
In contrast to conventional evaluation settings in
which separate policies are trained for each task, each training run in our
setting produces a single generalist policy that is evaluated on multiple
tasks. This results in correlated scores across tasks, and we hence treat independently trained policies, rather than task-policy pairs, as the
statistical units. Moreover, we deliberately use the arithmetic mean to
aggregate performance across tasks, rather than a robust statistic such as the median or IQM~\citep{agarwal2021Deep}.
This ensures that potentially poor performance on difficult tasks is not discarded as outliers.
Additionally, we provide per-task performance scores for fine-grained comparisons.
We report two-sided 95\% Student-$t$ CIs~\citep{student1908probable} over the
resulting policy-level estimates. A full description of our evaluation
protocol is provided in \cref{app:eval_protocol}.

\subsection{Environments and Evaluation Tasks}
\begin{figure*}[t]
   \centering
   \begin{subfigure}[t]{0.23\textwidth}
      \centering
      \includegraphics[width=.82\textwidth]{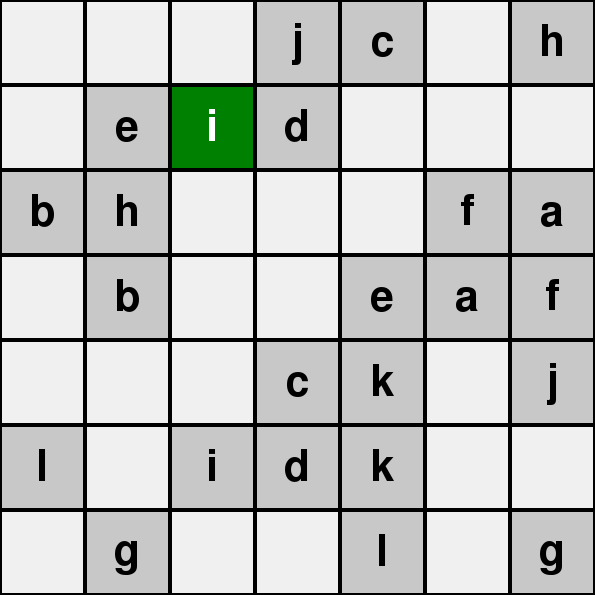}
      \caption{LetterWorld}
   \end{subfigure}\hfill
   \begin{subfigure}[t]{0.23\textwidth}
      \centering
      \includegraphics[width=\textwidth]{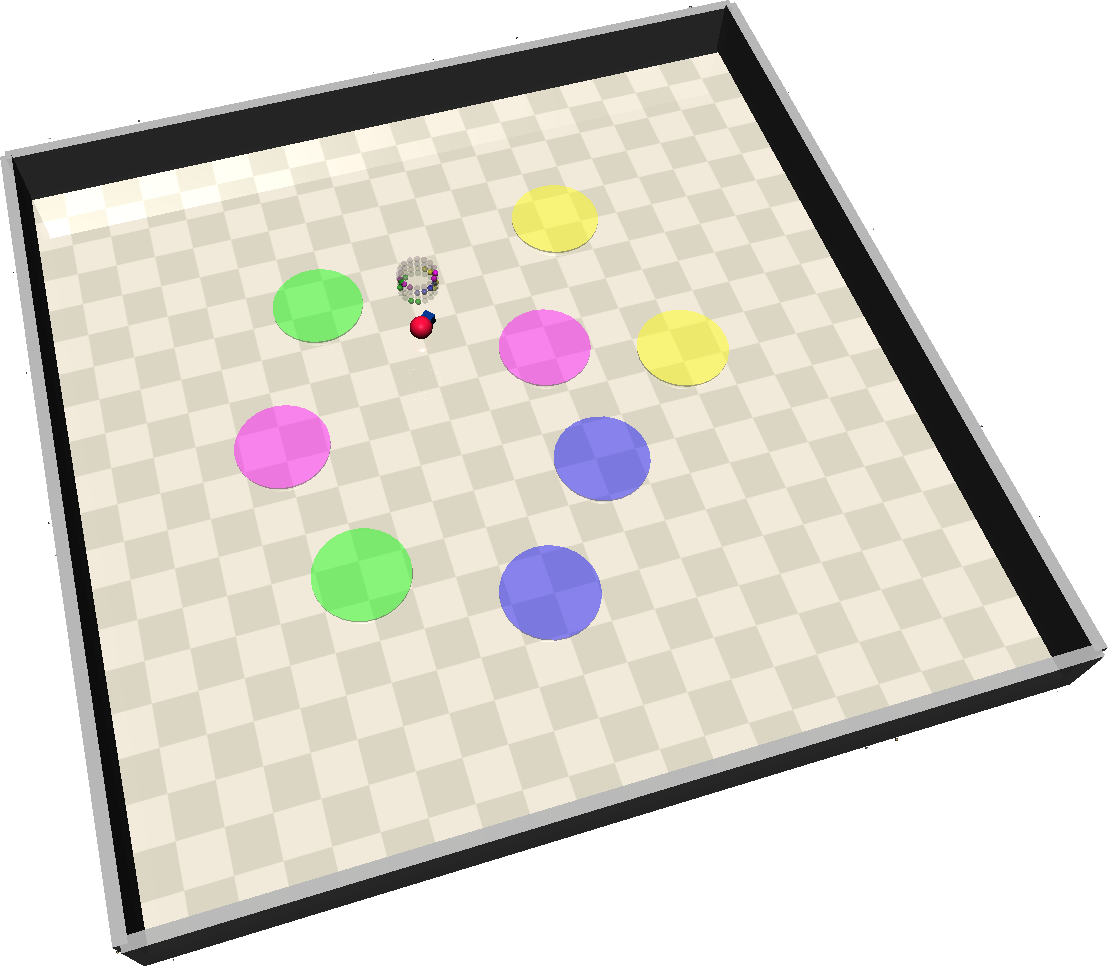}
      \caption{ZoneEnv}
   \end{subfigure}\hfill
   \begin{subfigure}[t]{0.23\textwidth}
      \centering
      \includegraphics[width=\textwidth]{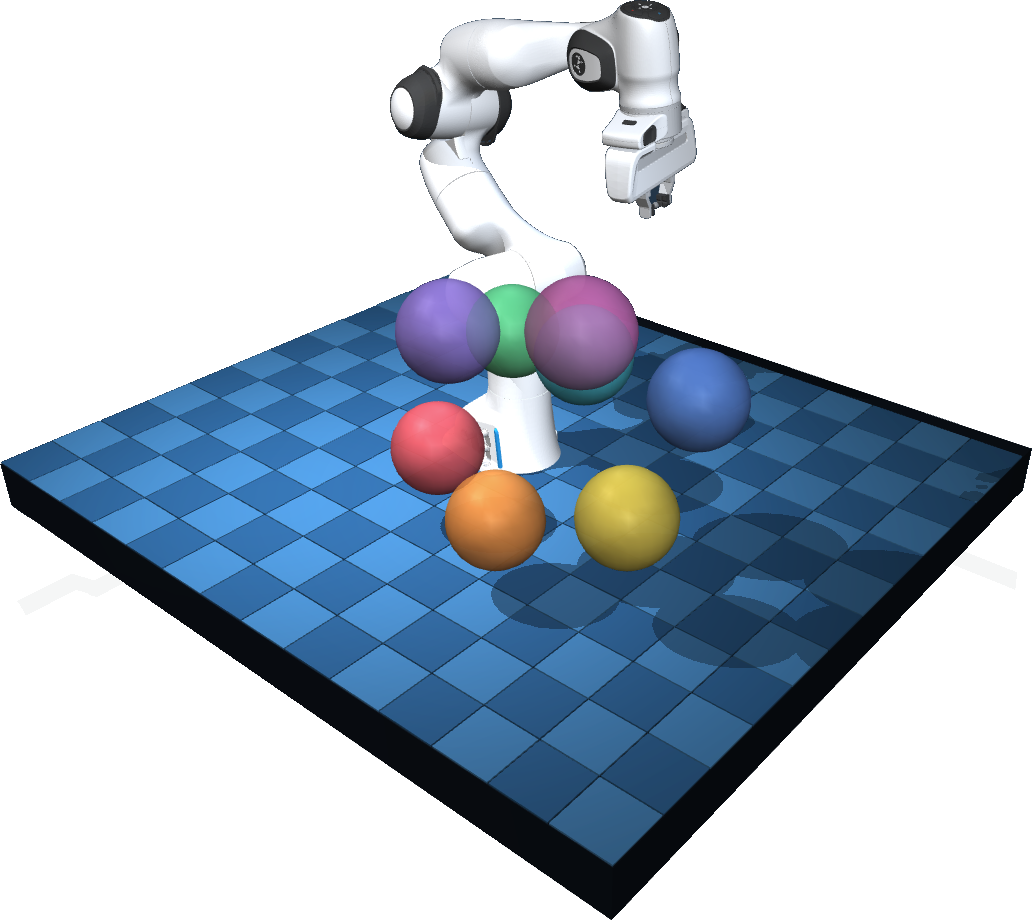}
      \caption{FrankaZoneEnv}
   \end{subfigure}\hfill
   \begin{subfigure}[t]{0.23\textwidth}
      \centering
      \includegraphics[width=\textwidth]{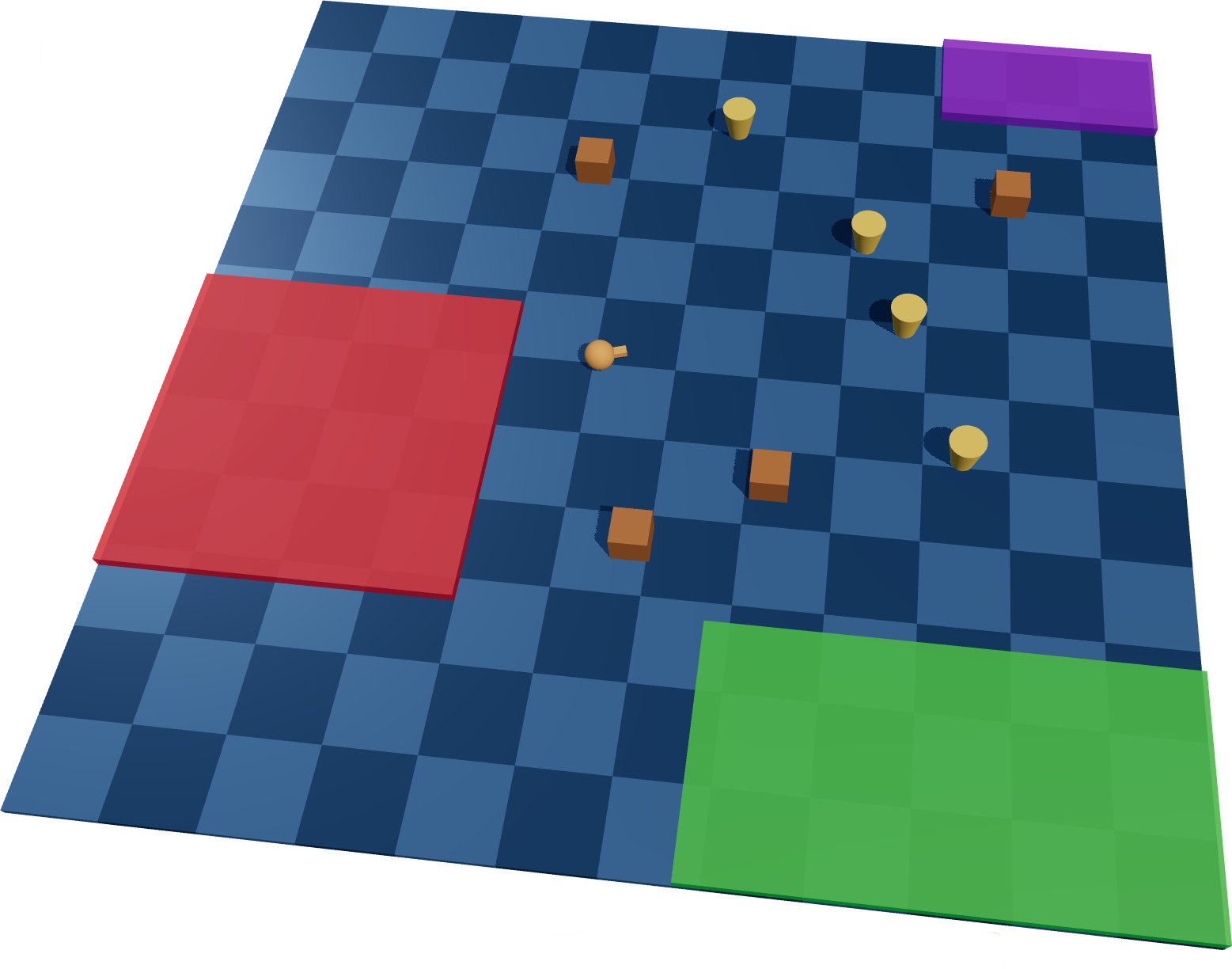}
      \caption{Warehouse}
   \end{subfigure}
   \caption{Visualisations of the main environments included in \jaxolotl.}
   \label{fig:envs}
\end{figure*}

\jaxolotl\ includes JAX implementations of four main environments spanning discrete navigation, continuous control, and object interaction, visualised in \cref{fig:envs}.
These are summarised below:
\begin{center}
\resizebox{.95\textwidth}{!}{%
{
\footnotesize
\begin{tabular}{@{}llll@{}}
   \toprule
   \textbf{Environment} & \textbf{Observation Space}                   & \textbf{Action Space}   & \textbf{Reference}              \\
   \midrule
   LetterWorld          & Grid ($7 \times 7 \times 13$)                & Discrete (4 directions) & \citet{vaezipoor2021LTL2Action} \\
   ZoneEnv              & {Proprioception} + LiDAR (69D) & Continuous (2D)         & \citet{vaezipoor2021LTL2Action} \\
   FrankaZoneEnv        & Proprioception\ + range--bearing (72D)             & Continuous (6D)         & \citet{guo2026SpecRLBench}      \\
   Warehouse            & Proprio.\ + LiDAR + inventory (47D)          & Hybrid (2D + 5D)        & \citet{jackermeier2026ZeroShot} \\
   \bottomrule
\end{tabular}%
}
}
\end{center}
LetterWorld, ZoneEnv, and Warehouse are implemented directly in JAX, whereas our FrankaZoneEnv implementation is based on MuJoCo XLA (MJX; \citealp{todorov2012mujoco}).
Each environment comes with curated finite- and infinite-horizon specification suites derived from tasks in the literature.
We provide further details on the environments and task suites in \cref{app:envs_task_suites}.

\subsection{Algorithms}
\label{sec:algs}
We provide unified implementations of six state-of-the-art multi-task LTL-RL methods:

\begin{center}
\resizebox{.95\textwidth}{!}{%
{
\footnotesize
\begin{tabular}{@{\extracolsep{\fill}}llcccl@{}}
   \toprule
   \textbf{Algorithm} & \textbf{Task representation} & \textbf{Full obs.} & \textbf{Infinite} & \textbf{Curriculum} & \textbf{Reference}              \\
   \midrule
   LTL2Action         & Formula syntax tree          & \tikzcmark      & \tikzxmark        & \tikzcmark          & \citet{vaezipoor2021LTL2Action} \\
   GCRL-LTL           & Proposition subgoal          & \tikzcmark      & \tikzcmark        & \tikzxmark          & \citet{qiu2023Instructing}      \\
   DeepLTL            & Reach--avoid sequence        & \tikzcmark      & \tikzcmark        & \tikzcmark          & \citet{jackermeier2025DeepLTL}  \\
   GenZ-LTL           & Reach--avoid subgoal         & \tikzxmark      & \tikzcmark        & \tikzxmark          & \citet{guo2025One}              \\
   SemLTL             & Semantic LDBA                & \tikzcmark      & \tikzcmark        & \tikzcmark          & \citet{abate2026Semantically}   \\
   StructLTL          & Boolean formula sequence     & \tikzcmark      & \tikzcmark        & \tikzcmark          & \citet{jackermeier2026ZeroShot} \\
   \bottomrule
\end{tabular}%
}
}
\end{center}

In contrast to the other methods, GenZ-LTL relies on domain-specific observation reduction functions (see \S\,\ref{sec:benchmark}) and LTL2Action does not support infinite-horizon tasks.
Our reusable curriculum learning implementation can be applied to any method, but the subgoal-based representations of GCRL-LTL and GenZ-LTL do not naturally allow for gradual increases in task complexity, and we hence do not apply curriculum learning (following the original works).
See \cref{app:methods} for details on methods.

\section{Experimental Evaluation}
Our experimental evaluation is split into two parts.
We first demonstrate the efficiency and correctness of \jaxolotl\ with respect to reference implementations (\S\,\ref{sec:efficiency}).
We then systematically evaluate existing methods on our benchmark suites to answer three research questions (\S\,\ref{sec:benchmark}):
\vspace{-0.5em}
\begin{enumerate}[label=\bfseries(Q\arabic*), itemsep=0.1em]
   \item How do existing approaches compare in terms of general task satisfaction on the curated finite- and infinite-horizon LTL suites?
   \item What are the capabilities of existing approaches for non-myopic reasoning?
   \item How do existing algorithms scale with the number of atomic propositions?
\end{enumerate}

\subsection{Efficiency and Correctness}
\label{sec:efficiency}

\paragraph{Computational Efficiency.}

\begin{figure}
   \centering
   \includegraphics[width=\textwidth]{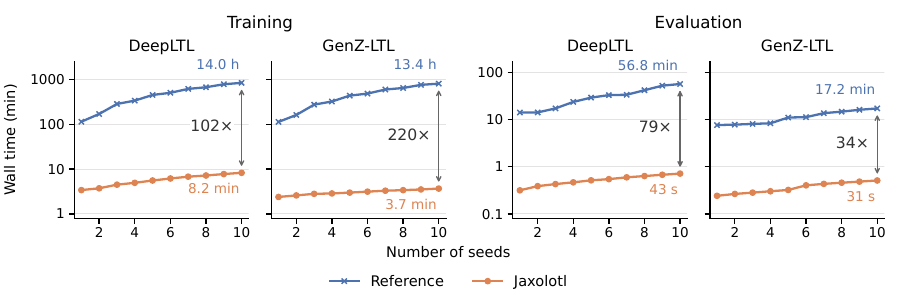}
   \caption{End-to-end training and evaluation times of \jaxolotl\ compared to original reference implementations on ZoneEnv. We achieve speedups of up to $220\times$ for training and $79\times$ for evaluation.}
   \label{fig:efficiency}
\end{figure}

We investigate the speed improvements enabled by \jaxolotl\ by comparing end-to-end training and evaluation times on ZoneEnv with original reference implementations.
We focus on DeepLTL and GenZ-LTL, two representative methods with distinct task representations.
All timings are measured on standard consumer hardware and include compilation overhead for the JAX implementations.
See \cref{app:computational} for full experimental details.

As shown in \cref{fig:efficiency}, our implementation yields speedups of $34\times$ and $47\times$ for training a single seed of DeepLTL and GenZ-LTL, respectively.
This difference becomes even more pronounced as we increase the number of independent runs, achieving efficiency improvements of $102\times$ and $220\times$ for 10 seeds.
This allows us to train 10 seeds of DeepLTL in 8.2 minutes and GenZ-LTL in 3.7 minutes, compared to 14.0 and 13.4 hours for the original implementations.
We further observe significant speedups for evaluation, with $79\times$ and $34\times$ improvements for DeepLTL and GenZ-LTL, respectively.
\cref{app:computational} provides additional comparisons on raw environment throughput.


\paragraph{Implementation Validation.}

\begin{figure}
   \centering
   \includegraphics[width=.9\linewidth]{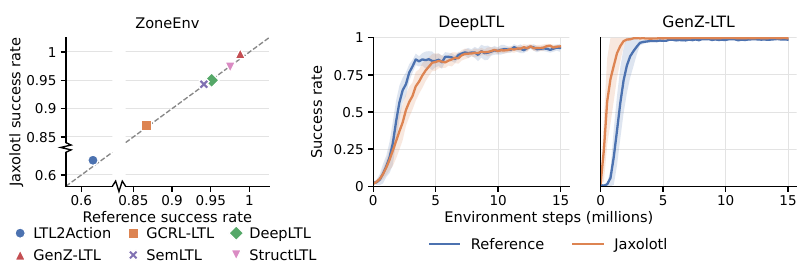}
   \caption{Validation of \jaxolotl. (Left) Final ZoneEnv success rates are within $\pm 1.1\%$ of the reference implementations for all methods. (Right) Overall training dynamics of DeepLTL and GenZ-LTL closely follow the references.}
   \label{fig:correctness}
\end{figure}

To validate the correctness of our implementations, we compare the performance of \jaxolotl\ against original reference implementations on ZoneEnv.
\cref{fig:correctness} compares final success rates across all methods (left), and training curves for DeepLTL and GenZ-LTL~(right).
The results demonstrate that our implementations closely match the references.
While there are small differences in the training curves, these can be explained by minor low-level implementation details.
Crucially, both overall learning behaviour and final success rates accurately reproduce the reference results.
Further experimental details are provided in \cref{app:correctness}.


\subsection{Benchmarking Existing Methods}
\label{sec:benchmark}

We proceed to systematically evaluate existing methods with \jaxolotl.
Detailed training parameters are described in \cref{app:training_details}.
The evaluation tasks in our general benchmark suite are drawn from the literature and enriched with curated specifications that cover a wide range of environment-specific behaviour and LTL objectives; see \cref{app:specs} for details.
Unless stated otherwise, we train each method using 10 independent random seeds, compute individual policy--task results over 512 episodes, and report $95\%$ Student-$t$ CIs over independent policies (see \refp{sec:eval_infra}).


\paragraph{GenZ-LTL Observation Interfaces.}
Standard GenZ-LTL uses domain-specific observation-reduction functions
that map the current observation $s_t$ into a task-relative representation,
containing only features relevant for the current reach--avoid subgoal
(see \cref{app:methods}).
We denote this configuration by GenZ-LTL$^\dagger$ throughout figures and
tables, and additionally evaluate \emph{GenZ-LTL (unreduced obs.)}, which conditions on the full observation together with a one-hot encoding of the
current subgoal, following \citet{guo2025One}.
The standard reduction assumes that observations decompose into
proposition-specific components and relies on a hand-designed fusion operator to produce fixed reach and avoid features.
The unreduced variant removes these assumptions while retaining the rest of the method's design, and is used when the assumptions do not hold, such as in Warehouse.

\subsubsection{General LTL Task Satisfaction \textbf{(Q1)}}
\label{sec:general}

\begin{figure}
   \centering
   \includegraphics[width=\linewidth]{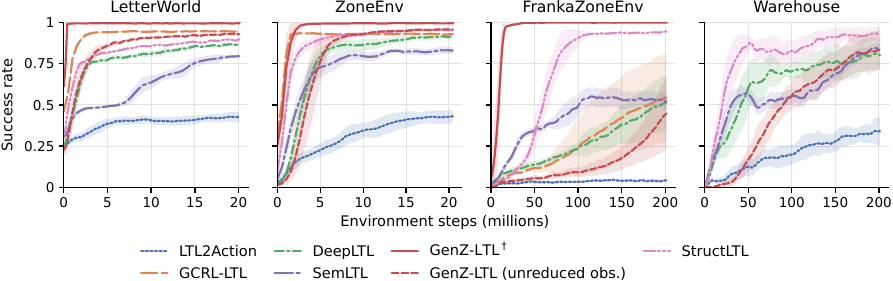}
   \caption{Aggregate finite-horizon success rates over training. Lines show means across 10 independently trained policies, with shaded $95\%$ Student-$t$ confidence intervals.
   }
   \label{fig:general_curves}
\end{figure}

\begingroup
\newcommand{\score}[2]{#1\ensuremath{_{\pm #2}}}
\newcommand{\best}[2]{\score{\textbf{#1}}{#2}}
\newcommand{\second}[2]{\score{\underline{#1}}{#2}}
\begin{table}[t]
       \centering
       \caption{Aggregated benchmark results over the \jaxolotl\ task suites, averaged
                over 512 episodes per formula, with $95\%$ CIs over 10 training seeds.
              Fin.: finite-horizon success rate (0--1);
              Inf.: completed accepting cycles with fixed episode length.
             R-S: completed accepting cycles on the reach-stay task suite.
             Best results are \textbf{bold}, second best \underline{underlined}, --- denotes unsupported configurations. See \cref{app:additional_resuls:per-spec_results} for per-specification results.}
       \label{tab:main-results}
       \footnotesize
       \setlength{\tabcolsep}{2.5pt}
       \resizebox{.95\textwidth}{!}{%
              \begin{tabular}{@{}l*{10}{c}@{}}
                     \toprule
                            & \multicolumn{2}{c}{LetterWorld}
                            & \multicolumn{3}{c}{ZoneEnv}
                            & \multicolumn{2}{c}{FrankaZoneEnv}
                            & \multicolumn{3}{c}{Warehouse}                             \\
                     \cmidrule(lr){2-3}\cmidrule(lr){4-6}\cmidrule(lr){7-8}\cmidrule(l){9-11}
                     Method & Fin.\,$\uparrow$                  & Inf.\,$\uparrow$
                            & Fin.\,$\uparrow$                  & Inf.\,$\uparrow$                  & R-S\,$\uparrow$
                            & Fin.\,$\uparrow$                  & Inf.\,$\uparrow$
                            & Fin.\,$\uparrow$                  & Inf.\,$\uparrow$                  & R-S\,$\uparrow$      \\
                     \midrule
                     LTL2Action
                            & \score{0.42}{0.03}             
                            & ---                            
                            & \score{0.43}{0.03}             
                            & ---                            
                            & ---                            
                            & \score{0.04}{0.01}             
                            & ---                            
                            & \score{0.34}{0.09}             
                            & ---                            
                            & ---                             \\
                     GCRL-LTL
                            & \second{0.94}{0.00}            
                            & \second{8.0}{0.1}              
                            & \score{0.93}{0.00}             
                            & \second{5.7}{0.1}              
                            & \score{30.2}{3.3}              
                            & \score{0.55}{0.27}             
                            & \score{8.1}{5.9}               
                            & ---                            
                            & ---                            
                            & ---                             \\
                     DeepLTL
                            & \score{0.87}{0.01}             
                            & \score{5.3}{0.2}               
                            & \score{0.92}{0.02}             
                            & \score{3.0}{0.3}               
                            & \second{540.1}{48.0}           
                            & \score{0.51}{0.17}             
                            & \score{4.8}{3.3}               
                            & \score{0.80}{0.09}             
                            & \second{2.6}{0.3}              
                            & \score{618.1}{118.7}            \\
                     GenZ-LTL$^\dagger$
                            & \best{0.99}{0.00}              
                            & \best{8.2}{0.0}                
                            & \best{1.00}{0.00}              
                            & \best{6.5}{0.2}                
                            & \score{492.0}{50.6}            
                            & \best{1.00}{0.00}              
                            & \best{35.0}{0.6}               
                            & ---                            
                            & ---                            
                            & ---                             \\
                     $\; \hookrightarrow$ unreduced obs.
                            & \score{0.93}{0.00}             
                            & \score{5.6}{0.0}               
                            & \second{0.96}{0.01}            
                            & \score{5.0}{0.1}               
                            & \score{321.6}{59.2}            
                            & \score{0.45}{0.22}             
                            & \score{5.1}{3.6}               
                            & \second{0.84}{0.12}            
                            & \score{2.2}{0.7}               
                            & \score{102.6}{24.8}             \\
                     SemLTL
                            & \score{0.79}{0.01}             
                            & \score{4.2}{0.1}               
                            & \score{0.83}{0.02}             
                            & \score{2.6}{0.1}               
                            & \score{532.9}{36.6}            
                            & \score{0.51}{0.05}             
                            & \score{13.6}{1.0}              
                            & \score{0.83}{0.04}             
                            & \score{0.2}{0.1}               
                            & \second{673.5}{30.9}            \\
                     StructLTL
                            & \score{0.90}{0.01}             
                            & \score{5.5}{0.2}               
                            & \score{0.95}{0.01}             
                            & \score{3.7}{0.3}               
                            & \best{572.7}{47.7}             
                            & \second{0.94}{0.00}            
                            & \second{15.3}{0.4}             
                            & \best{0.96}{0.01}              
                            & \best{3.1}{0.2}                
                            & \best{809.9}{30.4}              \\
                     \bottomrule
              \end{tabular}%
       }
\end{table}
\endgroup

\cref{tab:main-results} lists final aggregate results across all task suites, and \cref{fig:general_curves} shows training dynamics on \textbf{finite-horizon tasks}, which we focus on first.
We observe that GenZ-LTL performs exceptionally well when provided with a suitable observation reduction function, achieving near perfect success rates.
While its unreduced variant obtains high performance on LetterWorld and ZoneEnv, it struggles in more complex environments, where StructLTL dominates among the general methods.
GCRL-LTL, DeepLTL, and SemLTL overall perform similarly, but GCRL-LTL is not applicable to Warehouse due to its limited avoidance heuristic (see \cref{app:methods}).
The early method LTL2Action is consistently outperformed by more modern approaches, showing a distinct gap in success rates.

For \textbf{infinite-horizon tasks}, we report the established metric of average completed accepting cycles in the B\"uchi automaton constructed from the task~\citep{jackermeier2025DeepLTL,guo2026SpecRLBench}.
Performance on recurrence tasks (\textit{Inf.}\ suite) follows the same general trends as in the finite case.
We include reach-stay (i.e.\ persistence) tasks, which require the agent to infinitely maintain a proposition, for ZoneEnv and Warehouse, where this necessitates meaningful closed-loop behaviour.
Here, automata-based methods that account for non-determinism perform best, led by StructLTL.


\subsubsection{Non-Myopic Reasoning \textbf{(Q2)}}
\label{sec:non-myopic}

We first isolate non-myopic reasoning in \textit{ConveyorWorldSimple-$k$}, a simplified variant of \citet{abate2026Semantically}'s grid-based ConveyorWorld. At the first step, the agent enters one of two one-way conveyor belts, which it cannot leave, leading to rooms $A$ and $B$. It must then satisfy $\event (\mathsf{p}_1 \land \event (\mathsf{p}_2 \land \dots \land \event \mathsf{p}_{k}))$, where $\mathsf{p}_1, \dots, \mathsf{p}_{k-1}$ lie in order along both belts, but $\mathsf{p}_k$ is sampled uniformly from $\{\mathsf{a}, \mathsf{b}\}$, with $\mathsf{a}$ only reachable in $A$ and $\mathsf{b}$ only in $B$. The correct belt thus depends on a subgoal $k$ steps ahead, which a non-myopic method should be able to account for. Since the only observation is the agent's position, GenZ-LTL cannot use observation reduction here.

\begin{figure}
   \centering
   \includegraphics[width=.95\linewidth]{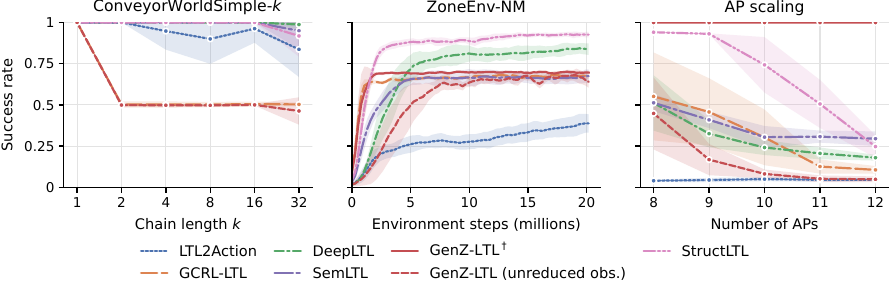}
   \caption{(Left) Non-myopic reasoning results on different step variants of ConveyorWorldSimple-$k$. (Middle) Aggregate finite-horizon success rates on ZoneEnv-NM over training. (Right) Results of scaling the number of atomic propositions in FrankaZoneEnv.}
   \label{fig:additional_results}
\end{figure}

\cref{fig:additional_results} (left) shows success rate over $k$. For $k>1$, GenZ-LTL and GCRL-LTL drop to $50\%$, the success rate of always entering the same belt. Both commit to a belt while pursuing $\mathsf{p}_1$, using an action policy conditioned only on the current subgoal, which is reachable via either belt. GCRL-LTL does select a complete subgoal sequence with its goal-conditioned value function (\cref{app:methods}), but its policy cannot distinguish between sequence-dependent ways of reaching the same subgoal.
The other methods, which condition on the full task, generally maintain high success rates up to $k=32$. Such long-horizon reasoning is much harder to learn once exploration and credit assignment become non-trivial (see original ConveyorWorld, \cref{app:additional_resuls:non_myopic_reasoning}).

To test non-myopic reasoning in a high-dimensional setting, we use a variation of ZonEnv (\textit{ZoneEnv-NM}, \citealp{jackermeier2026ZeroShot}), in which entering a purple zone removes all green zones for the rest of the episode, so the agent must avoid purple zones whenever a green zone is required later. On finite-horizon tasks (\cref{fig:additional_results}, middle), GenZ-LTL and GCRL-LTL plateau at $70.0\%$ and $66.9\%$, while StructLTL and DeepLTL reach $92.7\%$ and $83.6\%$. SemLTL plateaus at $67.5\%$ despite conditioning on the full task: this suggests that such \textit{conditioning} alone does not ensure non-myopic \textit{behaviour}, as the policy must still learn to exploit this information.
LTL2Action struggles to converge ($44.2\%$).

\subsubsection{Scaling with Atomic Propositions \textbf{(Q3)}}

We train each method in FrankaZoneEnv with an increasing number of propositions, evaluating on the same finite-horizon tasks (spanning the first $8$ propositions) throughout.
This isolates the difficulty of \textit{training} with an increasing number of propositions, while keeping the task difficulty fixed.

\cref{fig:additional_results} (right) presents the results. All methods except GenZ-LTL with observation reduction degrade as $|AP|$ grows. We attribute this mainly to grounding: each additional proposition must be associated with its own features in an increasingly complex observation space. This affects myopic methods too; GenZ-LTL (unreduced obs.) and GCRL-LTL lose most of their success rate by $|AP|=11$. Non-myopic methods must additionally generalise over a set of task sequences that grows exponentially with $|AP|$: StructLTL, the strongest general method up to $|AP|=9$, degrades sharply beyond that. GenZ-LTL is unaffected only because its hand-designed observation reduction maps zones to generic ``reach'' and ``avoid'' features, sidestepping the grounding problem by construction.




\section{Conclusion, Limitations, and Future Work}

We introduced \jaxolotl, a unified, high-performance benchmark suite for multi-task LTL-RL that enables controlled comparisons of existing methods at scale.
Our evaluation shows that existing methods have complementary strengths and limitations.
Methods that condition on the full task, such as StructLTL and DeepLTL, reason non-myopically, but, like all methods without observation reduction, they degrade as the number of atomic propositions grows.
GenZ-LTL scales by exploiting a useful prior in the form of observation reduction, but this requires manual engineering, is not always applicable, and remains myopic.
No evaluated method yet combines non-myopic reasoning, scaling to many propositions, and independence from environment-specific preprocessing.

\paragraph{Limitations.}
Our task suites are finite and curated by hand to capture interesting behaviour, but some of our results may depend on the specific task families that we used.
\jaxolotl\ currently only supports PPO training (as in the supported methods), although its abstractions are algorithm-agnostic.
Other restrictions stem from the methods rather than the framework: LTL2Action does not support infinite-horizon tasks, GCRL-LTL and GenZ-LTL with observation reduction cannot be applied to Warehouse, and all methods assume known propositions (for GenZ-LTL, a known space of propositions) and an accurate labelling function with access to privileged state information.

\paragraph{Future work.}
We plan to extend \jaxolotl\ with off-policy algorithms and additional environments, particularly robotics domains such as contact-rich manipulation, and to scale the benchmark to larger proposition vocabularies and richer task distributions.
For methods, the central open problems are relaxing the assumptions above, e.g.\ through learned or uncertain labelling functions and open-vocabulary settings, and designing a single approach with all three properties identified above.

\clearpage

\subsection*{AI use statement}

In this work, we used generative AI tools to implement methods: they helped
create and edit software code, including parts of the environment and method
implementations as well as plotting and tooling scripts. We have not used
generative AI tools to generate synthetic data, develop conceptual frameworks,
propose or refine hypotheses, design or give feedback on experiments, or
interpret results; formulating or proving mathematical claims, translation,
dataset cleaning and qualitative data analysis are not applicable to this work.
Additionally, we used generative AI tools to polish and proof read text
for readability and to find typos. We have reviewed all AI-assisted work:
an author reviewed every AI-assisted code change, the code is checked by
automated tests, and we validated our implementations against re-runs of the
official codebases and published results (\cref{app:correctness}). All AI-assisted
text was revised by the authors. We take responsibility for the final content
of this work, including text, claims or artifacts produced with the aid of
generative AI.




\subsection*{Reproducibility statement}

Our full implementation is available at \codeurl, including all environments, algorithms and task suites, together with pre-defined Hydra configurations for the entire benchmark suite. \Cref{app:envs_task_suites} describes the environments and task suites, \cref{app:methods} the implemented methods, and \cref{app:training_details} the model architectures, curricula and hyperparameters for each method and environment. \Cref{app:eval_protocol} formalises our evaluation protocol: unless stated otherwise, all results use 10 independent training runs, 512 evaluation episodes per policy and specification, and two-sided 95\% Student-$t$ confidence intervals over policies. \Cref{app:computational} specifies the hardware and timing protocol for the efficiency comparisons, \cref{app:correctness} the re-runs and reported results used to validate our implementations, and \cref{app:additional_resuls:per-spec_results} reports per-specification results for all benchmarks.

\subsubsection*{Acknowledgments}
This work was supported by the UKRI Erlangen AI Hub on Mathematical and Computational Foundations of AI (No. EP/Y028872/1). MJ and JC are funded by the EPSRC Centre for Doctoral Training in Autonomous Intelligent Machines and Systems (EP/S024050/1).

\bibliography{rl,mc,ml,jc}
\bibliographystyle{iclr2027_conference}

\clearpage
\appendix
\crefalias{section}{appendix}
\crefalias{subsection}{appendix}

\section{Extended Related Work}
\label{app:extended_related_work}

\paragraph{RL from Temporal Logic Specifications.}
A large body of work uses LTL to specify a single, fixed task to an RL agent.
The standard approach translates the formula into an automaton, typically a limit-deterministic B\"{u}chi automaton (LDBA; \citealp{sickert2016LimitDeterministic}), and learns a policy on the product of the MDP and the automaton \citep{sadigh2014Learning,hasanbeig2018LogicallyConstrained,hasanbeig2023Certified,hahn2019OmegaRegular,bozkurt2020Control}.
Subsequent work develops reward and discounting schemes with optimality guarantees \citep{voloshin2022Policy,voloshin2023Eventual,shao2023Sample}, improves exploration under the resulting sparse rewards \citep{bagatella2025Directed}, and characterises the fundamental limits of learning LTL objectives \citep{yang2022Intractability}.
Related work uses quantitative semantics of temporal logics as dense rewards \citep{aksaray2016QLearning,li2017Reinforcement,jothimurugan2019Composable}, or specifies tasks over finite traces \citep{camacho2019LTL,degiacomo2019Foundations}.
These methods learn a separate policy per specification, whereas we consider policies that generalise zero-shot to unseen specifications.

\paragraph{Reward Machines.}
Reward machines (RMs) expose the structure of a reward function as a finite-state machine \citep{toroicarte2018Using,toroicarte2022Reward}, enabling counterfactual experience sharing across machine states and automated reward shaping.
RMs can be constructed from formal specifications \citep{camacho2019LTL}, learned from experience \citep{toroicarte2019Learning,xu2020Joint}, composed hierarchically \citep{furelosblanco2023Hierarchies}, or grounded in raw observations via neural relaxations \citep{umili2024Neural}.
Since automata derived from LTL formulae can be viewed as RMs, many multi-task LTL-RL methods build on this perspective, although RM methods themselves typically target a single fixed machine.

\paragraph{Further LTL-Based Multi-Task RL Methods.}
\citet{toroicarte2018Teaching} learn options for co-safe LTL subformulae and use progression to transfer them to new tasks composed of known subformulae, while \citet{leon2021Systematic} study systematic generalisation to unseen formulae.
Other work encodes specifications with transformers \citep{zhang2023Exploiting,zhang2024Exploiting}, conditions policies on pretrained embeddings of compositional DFAs \citep{yalcinkaya2024Compositional}, learns future-dependent options \citep{xu2024Generalization}, or studies inductive generalisation from specifications \citep{subramanian2025Inductive}.
Most of these methods, like those we implement, assume access to a ground-truth labelling function mapping observations to propositions; \citet{pannacci2026Grounding} relax this by jointly learning a symbol grounder alongside the policy.

\paragraph{Instruction Following and Multi-Task RL.}
Beyond formal languages, instruction-following agents have been conditioned on natural language, policy sketches, and subtask graphs \citep{oh2017Zeroshot,andreas2017Modular,sohn2018Hierarchical,chevalierboisvert2019BabyAI}; see \citet{luketina2019Survey} for a survey.
Natural language is flexible but ambiguous, whereas LTL provides precise semantics and automata that track task progress.
Translating natural language into LTL is an active area of research \citep{cosler2023nl2spec,fuggitti2023NL2LTL,liu2023Lang2LTL}, suggesting that LTL-conditioned policies can serve as a backbone for language-driven agents.

\paragraph{Offline and Generative Approaches.}
A complementary line of work satisfies temporal logic specifications with generative models trained on offline data, including diffusion-based planners guided by LTL \citep{feng2024LTLDoGa, zoellner2026hint2}, flow matching over graph-encoded specifications \citep{meng2025TeLoGraF}, and specification-conditioned decision transformers \citep{guo2024Temporal}.
In robotic manipulation, \texttt{ManiGuard} \citep{peng2026ManiGuard} evaluates vision-language-action (VLA) policies on LTL\textsubscript{f} tasks.
These settings assume access to demonstration data or pretrained models, whereas \jaxolotl\ targets online RL.

\paragraph{RL Software.}
General-purpose libraries such as Gymnasium \citep{towers2025Gymnasium}, \textsc{Stable-Baselines3} \citep{raffin2021Stable}, CleanRL \citep{huang2022CleanRL}, and \texttt{skrl} \citep{antonio2023skrl}, as well as safe RL libraries such as Safety-Gymnasium \citep{ji2023Safety} and \texttt{OmniSafe} \citep{ji2024Omnisafe}, do not natively represent temporally extended task specifications.
LTL-based RL codebases therefore typically implement automaton construction and progression symbolically in Python on top of these libraries, which runs on the CPU and prevents end-to-end JIT compilation.
In addition to the JAX frameworks discussed in \cref{sec:related_work}, the ecosystem includes Pgx \citep{koyamada2023Pgx}, Kinetix \citep{matthews2025Kinetix}, Stoix \citep{toledo2024Stoix}, and JaxUED \citep{coward2024JaxUED}, none of which support LTL-specified tasks.

\paragraph{Evaluation Methodology.}
Concerns about the reliability of empirical RL results are long-standing \citep{henderson2018Deep,colas2018How}.
Recommended remedies include performance profiles and aggregate metrics with interval estimates \citep{jordan2020Evaluating,agarwal2021Deep}, careful experimental design \citep{patterson2024Empirical}, and community-wide standardised protocols \citep{gorsane2022Standardised,jordan2024Position}.
\jaxolotl\ adopts these recommendations, and its computational efficiency makes the required number of seeds and evaluation episodes practical.

\section{Semantics of LTL}
\label{app:ltl-semantics}

LTL semantics are defined over infinite words $\sigma\in(2^{AP})^\omega$.
The satisfaction relation $\sigma\models\varphi$ is defined recursively as~\citep{pnueli1977temporal}:
\begin{align*}
   \sigma & \models \top                                                                                                                                                                 \\
   \sigma & \models \mathsf a                        &  & \text{iff } \mathsf a\in \sigma_0                                                                                              \\
   \sigma & \models \varphi_1\land\varphi_2          &  & \text{iff } \sigma\models\varphi_1 \land \sigma\models\varphi_2                                                                \\
   \sigma & \models \neg\varphi                      &  & \text{iff } \sigma\not\models\varphi                                                                                           \\
   \sigma & \models \nex\varphi                      &  & \text{iff } \sigma_{1\ldots}\models\varphi                                                                                     \\
   \sigma & \models \varphi_1\;\mathsf{U}\;\varphi_2 &  & \text{iff } \exists j\geq 0.\; \sigma_{j\ldots}\models\varphi_2 \land \forall 0\leq i < j.\; \sigma_{i\ldots}\models\varphi_1.
\end{align*}

\section{Evaluation Protocol}
\label{app:eval_protocol}

This appendix formalises the evaluation protocol summarised in \cref{sec:eval_infra}, which adapts the evaluation principles of \citet{agarwal2021Deep} to generalist policies evaluated on a fixed set of benchmark tasks. Since each training run yields a single policy that is evaluated on every task, we treat independently trained policies, rather than policy–task pairs, as the statistical units.

Concretely, let $\Phi = \Phi_{\text{fin}} \uplus \Phi_\infty$ denote a fixed
set of finite- and infinite-horizon benchmark tasks for a given environment,
let $S$ denote the number of independently
trained policies (i.e.\ seeds), and let $E$ be the number of evaluation
episodes per policy and specification.
We first estimate
\begin{equation*}
   \hat{x}_{s,\varphi}
   = \frac{1}{E}\sum_{e=1}^{E} m_\varphi(\tau_{s,\varphi,e}),
\end{equation*}
where $m_\varphi$ is the evaluation metric for $\varphi$.
For finite-horizon specifications,
$m_\varphi(\tau) \coloneq \mathbb{1}[\tau\models\varphi]$ directly measures
satisfaction; for infinite-horizon specifications we use the number of visits to accepting states.
Since the initial state of every evaluation episode is drawn from the environment's episode initialisation distribution (\cref{app:envs}), $\hat{x}_{s,\varphi}$ is an unbiased estimate of the expected performance of policy $\pi_s$ on $\varphi$ under this distribution.

For a nonempty subset $\Psi \subseteq \Phi$ of specifications with comparable
metrics (e.g.\ all finite-horizon tasks), we define the per-policy aggregate
$\hat{x}_{s,\Psi} \coloneq
   |\Psi|^{-1}\sum_{\varphi\in\Psi}\hat{x}_{s,\varphi}$.
The overall point estimates across all evaluated runs are
\begin{equation*}
   \hat{\mu}_\varphi
   \coloneq \frac{1}{S}\sum_{s=1}^{S}\hat{x}_{s,\varphi},
   \qquad
   \hat{\mu}_\Psi
   \coloneq \frac{1}{S}\sum_{s=1}^{S}\hat{x}_{s,\Psi}.
\end{equation*}
Aggregating task scores within each policy before estimating uncertainty
preserves the correlations between scores obtained from the same training run.

To account for uncertainty across seeds, we report two-sided 95\%
Student-$t$ confidence intervals. For either an individual specification or
a task-set aggregate, let $z_s$ denote the corresponding policy-level
estimate, and define
\begin{equation*}
   \bar{z} \coloneq \frac{1}{S}\sum_{s=1}^{S}z_s,
   \qquad
   \hat{\sigma}_z^2
   \coloneq \frac{1}{S-1}\sum_{s=1}^{S}(z_s-\bar{z})^2.
\end{equation*}
The confidence interval is then
\begin{equation*}
   \left[
      \bar{z} - t_{S-1,0.975}\frac{\hat{\sigma}_z}{\sqrt{S}},
      \ \bar{z} + t_{S-1,0.975}\frac{\hat{\sigma}_z}{\sqrt{S}}
      \right],
\end{equation*}
where $t_{\nu,q}$ denotes the $q$-quantile of the Student-$t$ distribution
with $\nu$ degrees of freedom. We use $S=10$ independent training runs.
These intervals are exact for normally distributed policy-level estimates
and otherwise approximate; they quantify uncertainty under repeated training
and evaluation, conditional on the fixed benchmark tasks.
For performance curves, we apply the same procedure at each training
checkpoint, obtaining pointwise rather than simultaneous confidence intervals.

\section{Environments and Task Suites}
\label{app:envs_task_suites}

\subsection{Environments}
\label{app:envs}
We provide a brief overview of the four main environments supported in $\jaxolotl$.

\paragraph{LetterWorld~\citep{vaezipoor2021LTL2Action}.}
LetterWorld is a discrete navigation environment on a $7 \times 7$ grid on which 12 distinct letters are scattered, each appearing in two randomly chosen cells. The letters constitute the atomic propositions: a proposition holds exactly when the agent occupies a cell containing the corresponding letter, so every reachable assignment is a singleton set or the empty set. The agent starts each episode in a fixed corner cell and moves in the four cardinal directions via a discrete action space; the grid is toroidal, so moving across an edge wraps the agent around to the opposite side. The observation is an egocentric view of the entire grid, centred on the agent, encoded as one-hot letter channels plus an additional agent channel. Letter placements are resampled at the beginning of every episode, rejecting layouts in which some cell cannot be reached without crossing a letter; this guarantees that tasks requiring the avoidance of certain letters remain solvable.

\paragraph{ZoneEnv~\citep{vaezipoor2021LTL2Action}.}
ZoneEnv consists of a bounded $6.6 \times 6.6\,\mathrm{m}$ plane containing eight non-overlapping circular zones of radius $0.4\,\mathrm{m}$, two of each of four colours. The colours form the atomic propositions: a proposition holds while the agent is inside a zone of the corresponding colour, and since zones never overlap, all reachable assignments are singletons or empty. The agent is a point robot that applies a forward force along its current heading and controls its angular velocity through a 2-dimensional continuous action space. Observations combine proprioceptive information (acceleration, velocity, and angular velocity) with a per-colour LiDAR that partitions the agent's surroundings into 16 evenly spaced angular bins and reports an exponentially decaying signal based on the distance to the nearest zone of each colour in each bin. Both the zone layout and the initial agent pose are randomly sampled within a central $5 \times 5\,\mathrm{m}$ spawn area at the start of every episode, with zone centres kept at least $1.1\,\mathrm{m}$ apart, and episodes terminate prematurely if the agent leaves the arena. Our implementation replaces the MuJoCo-based point robot of the original environment with lightweight point-mass dynamics that retain the original control interface.

\paragraph{FrankaZoneEnv~\citep{guo2026SpecRLBench}.}
FrankaZoneEnv is a 3D robotic manipulation environment in which a 7-DoF Franka Panda arm, simulated with MuJoCo XLA (MJX; \citealp{todorov2012mujoco}), must visit coloured zones with its end effector. The zones are virtual spheres of radius $7\,\mathrm{cm}$ without contact geometry, one per colour (eight by default), placed uniformly at random in a $0.52 \times 0.64 \times 0.48\,\mathrm{m}$ box within the arm's reachable workspace; rejection sampling ensures that the sphere centres are at least $14\,\mathrm{cm}$ apart, reachable, and clear of the initial end-effector position. Each colour is an atomic proposition that holds while the grasp centre lies inside the corresponding sphere, so all reachable assignments are singletons or empty. The 6-dimensional continuous action specifies Cartesian pose increments (up to $2\,\mathrm{cm}$ of translation and $0.1\,\mathrm{rad}$ of rotation per step) of the end effector, which a differential inverse-kinematics controller converts into joint-space position targets. Observations consist of arm proprioception (joint angles and velocities, the actual and commanded end-effector poses, and an inverse-kinematics feasibility flag) together with range--bearing information for every zone, i.e.\ the relative position of and distance to each sphere. The initial arm configuration is perturbed around a home pose with up to $\pm 0.15\,\mathrm{rad}$ of per-joint noise, and the zone centres are resampled at the beginning of each episode. The number of colours is configurable, which makes the environment particularly suited to studying how methods scale with the number of propositions.

\paragraph{Warehouse~\citep{jackermeier2026ZeroShot}.}
The Warehouse environment couples continuous navigation with object interaction. A point robot with the same dynamics as in ZoneEnv moves in a bounded $6.6 \times 6.6\,\mathrm{m}$ world containing four vases and four crates at randomised positions, along with three fixed axis-aligned rectangular regions: region A ($2.2 \times 2.2\,\mathrm{m}$), region B ($2.9 \times 1.5\,\mathrm{m}$), and a door area ($1.6 \times 0.8\,\mathrm{m}$). The action space is hybrid: at every step, the agent applies a continuous (force, angular velocity) pair and simultaneously selects one of five discrete interactions (do nothing, pick up or drop a vase, pick up or drop a crate). An object can be picked up when the agent is within a radius of $0.2\,\mathrm{m}$ of it, and a carried object is deposited at the agent's current position when dropped, after which a 10-step cooldown prevents an immediate pickup. The five atomic propositions indicate whether the agent is inside each of the three regions and whether it is currently carrying a vase or a crate. In contrast to the other environments, multiple propositions can hold simultaneously (e.g.\ carrying both a vase and a crate while inside region A), giving rise to non-singleton assignments; since the regions are disjoint and at most one object of each type can be carried, 16 assignments are reachable in total. Observations comprise proprioception, the agent's global position, a per-object-type egocentric LiDAR with 16 angular bins, direction vectors towards the three regions, and binary carrying flags. Episodes terminate prematurely if the agent leaves the world.

\subsection{Task Suites}
\label{app:specs}

\cref{tab:ltl_specs_finite}, \cref{tab:ltl_specs_infinite}, and \cref{tab:ltl_specs_reach_stay} present the finite-horizon, infinite-horizon, and reach-stay LTL specifications used for evaluation in \cref{sec:general}, respectively. \cref{tab:ltl_specs_finite} also includes the specifications used to compute the evaluation curves for ZoneEnv-NM in \cref{sec:non-myopic}. For LetterWorld, ZoneEnv, Warehouse, and ZoneEnv-NM, these specifications were largely taken from the literature, and augmented with a broader coverage of more challenging tasks. For FrankaZoneEnv, the specifications are new and span a broad range of complex and interesting temporally extended behaviours; the pool of eight unique propositions allows for much greater compositional complexity compared to, e.g., ZoneEnv (which only names four unique propositions in its standard formula set).

As stated in \cref{sec:non-myopic}, ConveyorWorld and ConveyorWorldSimple evaluate on sequential reachability specifications of the form $(\mathsf{p}_1 \land \event (\mathsf{p}_2 \land \dots \land \event \mathsf{p}_{k}))$ where $\mathsf{p}_i \neq \mathsf{p}_j$, and propositions up to $\mathsf{p}_{k-1}$ are accessible in both rooms, but $\mathsf{p}_k \sim U(\{\mathsf{a}, \mathsf{b}\})$ is sampled evenly between $\mathsf{a}$ (only accessible in room $A$) and $\mathsf{b}$ (only accessible in room $B$).

\section{Supported LTL-RL Algorithms}
\label{app:methods}
We provide a brief characterisation of the multi-task LTL-RL methods supported in $\jaxolotl$, focusing on their task representations, task encoder architectures, and evaluation-time procedures. The corresponding training and model hyperparameters are listed in Appendix~\ref{app:hyperparameters}.

\paragraph{LTL2Action~\citep{vaezipoor2021LTL2Action}.}
LTL2Action conditions the policy directly on the syntax tree of the LTL formula and keeps the task Markovian via \emph{formula progression} \citep{bacchus2000Using}: after every environment step, the formula is rewritten into an equivalent residual formula expressing what remains to be satisfied given the observed assignment, and the task counts as satisfied (or violated) once the formula progresses to $\top$ (or $\bot$). The syntax tree of the current formula is encoded with a relational graph convolutional network (RGCN; \citealp{schlichtkrull2018Modeling}): every operator and proposition corresponds to a node with a learned embedding, and messages are passed from children to parents along three edge relations (unary operand, left and right binary operands) for a fixed number of rounds with shared weights, after which the embedding of the root node serves as the task embedding. Since messages only flow upwards, the number of message-passing rounds bounds the formula depth visible at the root, and must therefore be chosen to cover the deepest formulae in the task distribution. The policy is trained on randomly sampled reach--avoid formulae with a curriculum of increasing depth. Because progression only signals success for specifications that resolve to $\top$ after finitely many steps, LTL2Action does not support infinite-horizon tasks. In our implementation, the closure of each formula under progression is precomputed offline, reducing run-time progression to indexing a static transition table; at evaluation time, a new specification requires no further machinery, as the same progression mechanism applies unchanged.

\paragraph{GCRL-LTL~\citep{qiu2023Instructing}.}
GCRL-LTL trains a goal-conditioned policy to reach individual atomic propositions, sampled uniformly at random and presented to the policy as a learned embedding of the goal proposition; no LTL formulae or curriculum are involved during policy training. Once the policy has been trained, a goal-conditioned value function (GCVF) is fitted in a second phase: it estimates the value of pursuing a new goal $g'$ from states encountered while travelling towards a goal $g$, and, following the official implementation, is trained by regressing onto the critic's value estimates along successful trajectories of the frozen policy.

At evaluation time, a given LTL specification is converted to an LDBA, and a depth-first search extracts candidate reach--avoid paths representing accepting runs from every automaton state, with accepting loops unrolled for infinite-horizon tasks. Whenever the automaton state changes, GCRL-LTL scores each candidate path by accumulating negative log-values along its subgoals---the first edge using the policy's critic, and subsequent edges using the GCVF---and directs the goal-conditioned policy towards the first subgoal of the best-scoring path; $\varepsilon$-transitions of the LDBA are taken automatically whenever the plan calls for them. Avoidance is handled heuristically: if the value estimate for an avoid proposition is sufficiently high, the action that the policy would greedily take towards it is masked out of the action distribution. Since this masking operates on discrete action logits, GCRL-LTL runs on discretised variants of the continuous-control environments.
The proposition-based avoidance heuristic furthermore does not translate to environments with more complex avoidance requirements, making GCRL-LTL not applicable to Warehouse.

\paragraph{DeepLTL~\citep{jackermeier2025DeepLTL}.}
DeepLTL conditions the policy on sequences of reach--avoid pairs of assignment sets, which represent accepting runs of an LDBA and inform the policy of the subgoals that remain until satisfaction. Training requires no LTL formulae: the policy learns on randomly sampled reach--avoid sequences under a curriculum of increasing depth, including sequences that begin with an $\varepsilon$-transition and repeat their final element to capture stability ($\event \always$-type) tasks. To embed a sequence, each assignment is mapped to a learned embedding (with a dedicated token for $\varepsilon$-transitions), the reach and avoid sets of every step are pooled by a permutation-invariant deep sets network \citep{zaheer2017Deep}, and the resulting per-step representations are processed by a gated recurrent unit (GRU; \citealp{cho2014Properties}) traversed from the end of the sequence backwards, so that its final hidden state is aligned with the current subgoal. The $\varepsilon$-transitions of the LDBA, which resolve its non-determinism, are exposed to the agent as an additional binary action that advances the automaton without affecting the environment.

At evaluation time, a given specification is converted into an LDBA, and a depth-first search enumerates the reach--avoid sequences corresponding to possible accepting runs from every automaton state, unrolling accepting loops for infinite-horizon tasks. Whenever the automaton state changes, the agent selects among the candidate sequences of the new state using its learned value function. Since the policy is trained on arbitrary sequences, it generalises across automata, and hence across LTL specifications.

\paragraph{GenZ-LTL~\citep{guo2025One}.}
GenZ-LTL is a decomposition-based method that handles one reach--avoid subgoal at a time, where a subgoal consists of a single assignment to reach and a set of assignments to avoid. Instead of learning a task embedding, GenZ-LTL relies on hand-designed \emph{observation reduction} functions that map the observation into a subgoal-relative form containing only the features relevant to the current subgoal, rendering the policy invariant to the number and identity of propositions. In ZoneEnv, for instance, the per-colour LiDARs are reduced to the LiDAR of the reach colour and an element-wise maximum over the LiDARs of the avoid colours; in LetterWorld, the per-letter grid channels are collapsed into a single channel marking reach and avoid cells; and in FrankaZoneEnv, the range--bearing features of the reach zone are retained alongside the avoid zones sorted by distance. This reduction assumes that observations decompose into proposition-specific components; where this does not hold, the unreduced variant conditions on the full observation together with one-hot encodings of the subgoal. The policy is trained on randomly sampled subgoals without a curriculum, using a safe-RL formulation: a Lagrangian-constrained variant of PPO in which a cost critic estimates the (maximum-based, reachability-style) probability of violating the avoid set, weighted against the reward objective by a state-dependent Lagrange-multiplier network.

At evaluation time, a specification is converted into an LDBA, and the first reach--avoid pair of every accepting run from the current automaton state yields the set of candidate subgoals. The agent selects the subgoal maximising the learned value minus the multiplier-weighted cost estimate, re-selects when the automaton state changes or a subgoal times out, and takes $\varepsilon$-transitions automatically whenever doing so is safe.

\paragraph{SemLTL~\citep{abate2026Semantically}.}
SemLTL conditions the policy on \emph{semantic state labels} of an LDBA. Each specification is compiled into a semantically labelled LDBA, in which every automaton state carries a feature vector summarising the semantics of its associated formulae; crucially, semantically equivalent states of different automata receive identical labels, which is what enables generalisation across specifications. The dimensionality of this label vector (the semantic input size) depends on the number of atomic propositions, and a single learned linear projection maps it to the task embedding. To resolve the non-determinism of the LDBA, the policy additionally receives the labels of all states reachable via $\varepsilon$-transitions, and a dedicated categorical action head chooses at every step between acting in the environment and jumping to one of the $\varepsilon$-successors. The policy is trained on sampled LTL formulae---including infinite-horizon $\always \event$- and $\event \always$-type tasks---under a curriculum of increasing complexity, receiving reward for accepting transitions of the automaton and a penalty for entering rejecting sink states. At evaluation time, a new specification is simply compiled into its semantically labelled LDBA and the policy acts directly on the resulting labels; no search or planning is involved.

\paragraph{StructLTL~\citep{jackermeier2026ZeroShot}.}
StructLTL conditions the policy on sequences of reach--avoid pairs of Boolean formulae over propositions, which represent the transitions along accepting runs of an LDBA: the reach formula is a conjunction of literals that characterises the transition to be taken, while the avoid formula is a disjunction of such conjunctions that must not be triggered. The encoder mirrors this logical structure hierarchically: propositions receive learned embeddings, with negation applied as a learned linear transform; a first deep sets module (the clause network) pools the literals of each conjunction; a second deep sets module (the disjunct network) pools the clause embeddings of each avoid formula; and the resulting per-step subgoal representations are combined by a single-head attention mechanism \citep{vaswani2017AttentionAllYou} in which the current subgoal attends over future ones, with a linear ALiBi position bias~\citep{press2021Train} whose slope controls how strongly attention is biased towards nearby subgoals. As in DeepLTL, training requires no LTL formulae: the policy learns on randomly sampled sequences of Boolean reach--avoid pairs under a curriculum, including $\varepsilon$-prefixed stability sequences, and $\varepsilon$-transitions are exposed as an additional binary action.

At evaluation time, a specification is converted into an LDBA and a depth-first search extracts accepting runs from every automaton state; the assignment sets labelling each transition are then synthesised into minimal Boolean formulae, with disjunctive reach formulae split into one candidate sequence per disjunct. Whenever the automaton state changes, the agent re-selects among the candidate sequences of the new state using its learned value function. Having been trained on arbitrary formula sequences, the policy generalises across automata and therefore across LTL specifications.

\section{Training Details}
\label{app:training_details}

\subsection{Model Architectures}
\label{app:model_architecture}

All methods in $\jaxolotl$ instantiate the same modular policy architecture, assembled from interchangeable components configured per environment. An observation encoder (``Env Net'' in the tables of \cref{app:hyperparameters}) maps the environment observation to a feature vector: in LetterWorld, a convolutional network processes the egocentric grid channels, whereas in every other environment the observation is flattened and passed through a multi-layer perceptron (MLP; \citealp{rumelhart1986Learning}). In parallel, the method-specific task encoder of \cref{app:methods} (``LTL Encoder'') produces an embedding of the current task representation. The observation features and task embedding are concatenated into a joint representation on which both the actor and the critic operate; the two encoders are thus shared, while the actor and critic themselves have no common parameters. The critic is an MLP mapping the joint representation to a scalar value estimate. The actor is an MLP whose final hidden layer feeds one linear head per parameter group of the action distribution. GenZ-LTL deviates from this pattern only in that it has no task encoder: its observation reduction (see Appendix~\ref{app:methods}) injects the subgoal into the observation itself, so the joint representation coincides with the encoded (reduced) observation; in the unreduced variant, one-hot encodings of the reach and avoid assignments are appended to the flattened observation before encoding. All linear layers are initialised with orthogonal weights and zero biases.

The actor heads follow the action space of the environment. In LetterWorld and ConveyorWorld, as well as in the discretised environment variants on which GCRL-LTL operates, a single head outputs the logits of a categorical distribution over the discrete actions. In ZoneEnv, ZoneEnv-NM, and FrankaZoneEnv, two heads output the mean and the state-dependent standard deviation of a diagonal Gaussian, with a softplus transformation guaranteeing positivity of the latter. In Warehouse, whose hybrid action space pairs a continuous control input with a discrete interaction, the Gaussian heads are complemented by a categorical head, and the resulting distributions are treated as independent factors of a joint distribution.

Methods that expose the $\varepsilon$-transitions of the LDBA to the agent extend the actor with dedicated heads. In DeepLTL and StructLTL, an additional scalar head defines a Bernoulli distribution over taking the $\varepsilon$-action, masked out whenever no $\varepsilon$-transition is available in the current automaton state; if the $\varepsilon$-action is taken, the sampled environment action is discarded and only the automaton advances. In SemLTL, the analogous head is instead evaluated once per $\varepsilon$-successor---on a joint representation formed from the observation features and the projected semantic label of that successor---and once on the current state, yielding a masked categorical distribution that selects between jumping to one of the $\varepsilon$-successors and acting in the environment.

Two methods maintain auxiliary networks alongside the actor and critic. GenZ-LTL, trained with the Lagrangian-constrained variant of PPO, adds a cost critic with the same structure as the critic, together with a state-dependent Lagrange-multiplier network whose softplus activation keeps the multiplier non-negative; both operate on the same encoded observation as the actor. GCRL-LTL fits its goal-conditioned value function as an entirely separate model comprising its own observation encoder and goal-embedding table: the encoded observation is concatenated with the embeddings of the currently pursued and the prospective goal propositions and mapped to a scalar value by an MLP.

\subsection{Curricula}
\label{app:curricula}

A curriculum in $\jaxolotl$ is a sequence of stages, each of which defines a distribution over tasks via a sampler (or a mixture of samplers with associated probabilities). Since sampled tasks are dynamic Python objects, curricula are precomputed: before training, we draw a fixed number of tasks per stage and convert them into the method-specific static array representation. At the beginning of each episode, a task is then sampled uniformly at random from the precomputed samples of the current stage. If sampling is expensive (as for SemLTL, which compiles every sampled formula into an LDBA), the precomputed curriculum is generated once and cached on disk.

The agent progresses through the stages based on a stage frontier shared across all parallel environments. We track the success rate at the frontier stage over a rolling window of recently completed episodes, where an episode counts as successful if it terminates with positive reward. The frontier advances once this rate exceeds the threshold of the current stage, provided that the window has been filled and that a minimum fraction of environments has contributed episodes at the frontier (this prevents a small number of environments from dominating the estimate). Environments behind the frontier do not switch immediately: after each training rollout, a lagging environment advances one stage with a fixed adoption probability, so that the training distribution shifts gradually.

The task distributions reflect the task representation of each method. DeepLTL and StructLTL are trained on reach--avoid sequences, where later stages increase the sequence depth and the number of reach and avoid propositions per step, and additionally include reach--stay tasks in environments with infinite-horizon specifications. For StructLTL, the reach and avoid components are Boolean formulae drawn from environment-specific pools rather than single assignments; the two methods share identical curricula in several environments. LTL2Action and SemLTL sample LTL formulae from templates of the corresponding families (sequenced reach and reach--avoid objectives, as well as reach--stay objectives for SemLTL), such that all curriculum-based methods are trained on comparable task distributions. In all cases, the samplers ensure that sampled tasks are feasible, e.g.\ by never sampling a reach objective that is already satisfied by the preceding step. As discussed in \cref{sec:algs}, GenZ-LTL and GCRL-LTL are trained on a single stage without progression, which samples uniformly random reach--avoid subgoals and goal propositions, respectively.

Depending on the method and environment, curricula consist of between one and seven stages. Where the original works prescribe curricula, ours closely follow them; for the remaining environments, we design analogous curricula. The exact stage definitions for each method and environment (samplers, parameter ranges, mixture weights, and thresholds) can be found in the accompanying codebase.

\subsection{Hyperparameters}
\label{app:hyperparameters}

We provide the full set of training and model hyperparameters for each method in each environment; see \cref{tab:hyperparameters_letter_world} for LetterWorld, \cref{tab:hyperparameters_zone_env} for ZoneEnv and ZoneEnv-NM, \cref{tab:hyperparameters_franka_zone_env} for FrankaZoneEnv, \cref{tab:hyperparameters_warehouse} for Warehouse, and \cref{tab:hyperparameters_conveyor} for ConveyorWorld and ConveyorWorldSimple. For GenZ-LTL and GCRL-LTL, the training ``curriculum'' consists of a single stage (i.e.\ no curriculum), so hyperparameters related to curriculum progression do not apply. Hyperparameters were largely taken from the literature where applicable. We use Adam~\citep{kingma2015Adam} for all experiments.

\section{Experimental Details}

\subsection{Computational Efficiency}
\label{app:computational}


All timing experiments were run on a single desktop machine with consumer hardware: an NVIDIA GeForce RTX 5070 Ti GPU (16\,GiB VRAM), an Intel Core i5-13600K CPU, and 32\,GiB of RAM. For the comparison in \cref{fig:efficiency}, \jaxolotl\ trains multiple seeds in parallel on the GPU by vectorising the training loop over seeds with \texttt{vmap}. The reference implementations train a single seed per run. Running two reference instances concurrently increases their overall throughput, whereas running more reduces it due to resource contention. For each number of seeds $k$, we therefore report the fastest wall-clock time in which each implementation produces $k$ trained seeds, and measure evaluation times in the same way. Timings for \jaxolotl\ include JIT compilation.

\begin{figure}
   \centering
   \includegraphics[width=\linewidth]{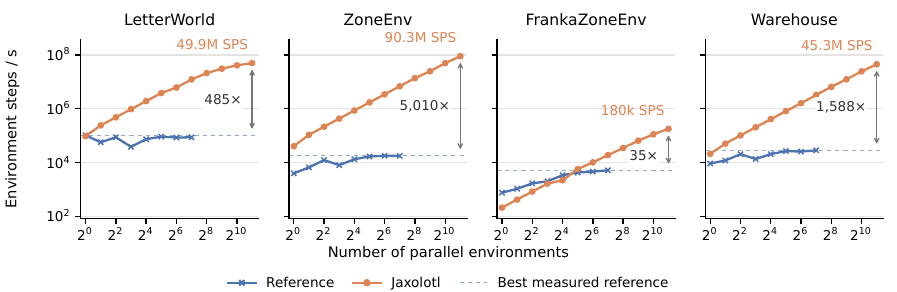}
   \caption{Environment throughput of \jaxolotl\ and the reference implementations against the number of parallel environments (log scales). The references run at most 128 parallel environments; the dashed line marks their best measured throughput. Annotations give the peak throughput of \jaxolotl\ and its speedup over this best reference measurement.}
   \label{fig:env_throughput}
\end{figure}

\paragraph{Environment Throughput.}
\Cref{fig:env_throughput} compares raw environment throughput, measured in environment steps per second (SPS) over 8,192 steps of each parallel environment. \jaxolotl\ batches environments on the GPU with \texttt{vmap}, whereas the reference implementations run each environment in a separate process and step them asynchronously. We evaluate \jaxolotl\ with up to 2,048 parallel environments and the references with at most 128, since larger numbers caused them to crash. At its largest setting, \jaxolotl\ is between 35$\times$ (FrankaZoneEnv) and 5,010$\times$ (ZoneEnv) faster than the best reference measurement. On FrankaZoneEnv, which simulates a 7-DoF arm with MJX, \jaxolotl\ only overtakes the reference from 32 parallel environments onwards. Part of the ZoneEnv speedup comes from replacing the MuJoCo-based point robot of the original environment with point-mass dynamics (\cref{app:envs}).


\subsection{Implementation Validation}
\label{app:correctness}


We validate our implementations on the finite-horizon ZoneEnv task set of \citet{jackermeier2025DeepLTL}. We re-run DeepLTL, GenZ-LTL, SemLTL and StructLTL with their official codebases over 10 seeds on this task set and compare them to our implementations on the same tasks. The reference success rates in \cref{fig:correctness} (left) come from these re-runs, except for LTL2Action and GCRL-LTL, for which we use the results reported by \citet{jackermeier2025DeepLTL} on the same task set. The training curves in \cref{fig:correctness} (right) show the re-runs of DeepLTL and GenZ-LTL.

\section{Additional Results}

\subsection{Per-Specification Results}
\label{app:additional_resuls:per-spec_results}

\cref{tab:appendix-results-finite}, \cref{tab:appendix-results-infinite}, and \cref{tab:appendix-results-reach-stay} present per-specification benchmark results (aggregated in \cref{sec:general}) for the finite-horizon, infinite-horizon, and reach-stay LTL specifications, respectively.

\subsection{Non-Myopic Reasoning}
\label{app:additional_resuls:non_myopic_reasoning}

\begin{figure*}[t]
   \centering
   \begin{subfigure}{0.38\textwidth}
      \centering
      \includegraphics[width=\textwidth]{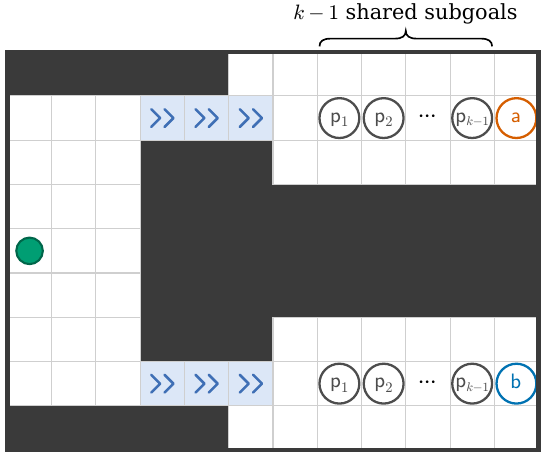}
      \caption{}
      \label{fig:non-myopia:conveyor:diagram}
   \end{subfigure}
   \hfill
   \begin{subfigure}{0.6\textwidth}
      \centering
      \includegraphics[width=\textwidth]{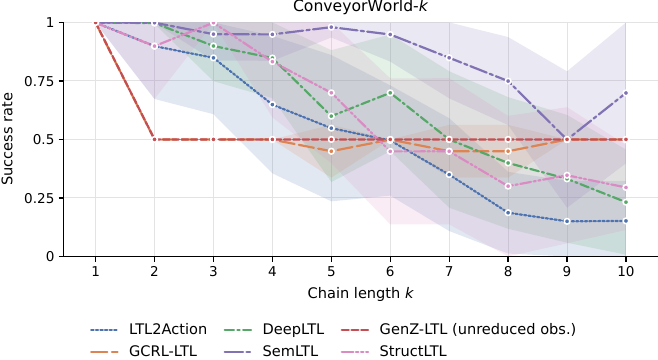}
      \caption{}
      \label{fig:non-myopia:conveyor:results}
   \end{subfigure}
   \caption{(a) A visualisation of ConveyorWorld-$k$; the green circle marks the agent start position, and the upper and lower one-way conveyor belts lead to rooms $A$ and $B$ respectively. (b) Non-myopic reasoning results on different step variants of ConveyorWorld-$k$.}
   \label{fig:non-myopia:conveyor}
\end{figure*}

While many of the methods studied are capable of non-myopic reasoning in principle, reasoning over many steps becomes difficult to learn in practice once exploration and credit assignment are non-trivial. To demonstrate, we evaluate them on $k$-step variants of the original \textit{ConveyorWorld} environment \citep{abate2026Semantically}, of which ConveyorWorldSimple-$k$ (\cref{sec:non-myopic}) is a simplification. The task structure is the same: the agent must commit to one of two one-way conveyor belts leading to rooms $A$ and $B$ before satisfying $\event (\mathsf{p}_1 \land \event (\mathsf{p}_2 \land \dots \land \event \mathsf{p}_{k}))$ with $\mathsf{p}_i \neq \mathsf{p}_j$, where $\mathsf{p}_1, \dots, \mathsf{p}_{k-1}$ are accessible in both rooms and $\mathsf{p}_k \sim U(\{\mathsf{a}, \mathsf{b}\})$ is accessible only in room $A$ (for $\mathsf{a}$) or $B$ (for $\mathsf{b}$). Unlike the simplified version, the agent must first explore the starting room to reach the belts, and then explore the room it arrives in to enter the grid tile associated with each proposition in order from $\mathsf{p}_1$ to $\mathsf{p}_k$ before receiving any reward. \cref{fig:non-myopia:conveyor:diagram} provides a visualisation of ConveyorWorld-$k$.

\cref{fig:non-myopia:conveyor:results} presents success rate over $k$ in ConveyorWorld; we see severe degradation of all non-myopic methods as $k$ increases. The performance of trained policies is \textit{trimodal} across seeds for each: some seeds learn the task and achieve $100\%$ success rate, others collapse to always entering the same room as a local optimum, and the rest never learn to exploit any reward within the training budget (e.g.\ continually staying in the initial room). The distribution generally shifts from the first mode through the second into the last with increasing $k$ as exploration and credit assignment become more challenging, since the agent must visit all $k$ subgoals in order before receiving any reward. Note that using a multi-stage curriculum did not improve performance, as many seeds failed to learn even simple $\event \mathsf{p}_k$ reach tasks in the first curriculum stage.

\begin{table}[ht]
\caption{Finite-horizon LTL specifications used for evaluation.}
\centering
\small
\label{tab:ltl_specs_finite}
\begin{tabular}{c l p{12cm}}
\toprule
 & {ID} & {LTL Formula} \\
\midrule
\multirow{10}{*}{\rotatebox[origin=c]{90}{LetterWorld}} & $\varphi_{1}$ & $\event (\mathsf{a} \wedge (\neg \mathsf{b} \until \mathsf{c})) \wedge \event \mathsf{d}$ \\
 & $\varphi_{2}$ & $(\event \mathsf{d}) \wedge (\neg \mathsf{f} \until (\mathsf{d} \wedge \event \mathsf{b}))$ \\
 & $\varphi_{3}$ & $(\event ((\mathsf{a} \vee \mathsf{c} \vee \mathsf{j}) \wedge \event \mathsf{b})) \wedge (\event (\mathsf{c} \wedge \event \mathsf{d})) \wedge \event \mathsf{k}$ \\
 & $\varphi_{4}$ & $\neg \mathsf{a} \until (\mathsf{b} \wedge (\neg \mathsf{c} \until (\mathsf{d} \wedge (\neg \mathsf{e} \until \mathsf{f}))))$ \\
 & $\varphi_{5}$ & $\always \neg (\mathsf{k} \vee \mathsf{l}) \wedge \event (\mathsf{a} \wedge \event (\mathsf{e} \wedge \event (\mathsf{i} \wedge \event \mathsf{d})))$ \\
 & $\varphi_{6}$ & $\always \neg (\mathsf{a} \vee \mathsf{b} \vee \mathsf{c} \vee \mathsf{d}) \wedge \always (\mathsf{h} \rightarrow \neg \mathsf{f} \until \mathsf{g}) \wedge \event \mathsf{g} \wedge \event (\mathsf{h} \wedge \event \mathsf{i})$ \\
 & $\varphi_{7}$ & $\event ((\mathsf{a} \vee \mathsf{l}) \wedge ((\mathsf{a} \rightarrow (\neg \mathsf{b} \until (\mathsf{c} \wedge \event \mathsf{d}))) \wedge (\mathsf{l} \rightarrow \event \mathsf{g})))$ \\
 & $\varphi_{8}$ & $\event (\mathsf{a} \wedge \event (\mathsf{b} \wedge \event (\mathsf{c} \wedge \event (\mathsf{d} \wedge \event (\mathsf{e} \wedge \event (\mathsf{f} \wedge \event (\mathsf{g} \wedge \event \mathsf{h})))))))$ \\
 & $\varphi_{9}$ & $\always \neg \mathsf{l} \wedge \event (\mathsf{i} \wedge \event (\mathsf{e} \wedge \event (\mathsf{a} \wedge \event (\mathsf{k} \wedge \event (\mathsf{g} \wedge \event \mathsf{c})))))$ \\
 & $\varphi_{10}$ & $\neg (\mathsf{b} \vee \mathsf{d} \vee \mathsf{f} \vee \mathsf{h} \vee \mathsf{j}) \until (\mathsf{l} \wedge \event ((\mathsf{a} \vee \mathsf{k}) \wedge ((\mathsf{a} \rightarrow \event \mathsf{c}))))$ \\
\midrule
\multirow{10}{*}{\rotatebox[origin=c]{90}{ZoneEnv}} & $\varphi_{11}$ & $\event (\mathsf{green} \wedge (\neg \mathsf{red} \until \mathsf{yellow})) \wedge \event \mathsf{purple}$ \\
 & $\varphi_{12}$ & $(\event \mathsf{red}) \wedge (\neg \mathsf{red} \until (\mathsf{green} \wedge \event \mathsf{yellow}))$ \\
 & $\varphi_{13}$ & $\event (\mathsf{red} \vee \mathsf{green}) \wedge \event \mathsf{yellow} \wedge \event \mathsf{purple}$ \\
 & $\varphi_{14}$ & $\neg (\mathsf{purple} \vee \mathsf{yellow}) \until (\mathsf{red} \wedge \event \mathsf{green})$ \\
 & $\varphi_{15}$ & $\neg \mathsf{green} \until ((\mathsf{red} \vee \mathsf{purple}) \wedge (\neg \mathsf{green} \until \mathsf{yellow}))$ \\
 & $\varphi_{16}$ & $((\mathsf{green} \vee \mathsf{red}) \rightarrow (\neg \mathsf{yellow} \until \mathsf{purple})) \until \mathsf{yellow}$ \\
 & $\varphi_{17}$ & $\always \neg \mathsf{purple} \wedge \event (\mathsf{red} \wedge \event (\mathsf{green} \wedge \event \mathsf{yellow}))$ \\
 & $\varphi_{18}$ & $\neg (\mathsf{green} \vee \mathsf{purple} \vee \mathsf{yellow}) \until (\mathsf{red} \wedge \event (\mathsf{yellow} \wedge (\neg \mathsf{red} \until \mathsf{green})))$ \\
 & $\varphi_{19}$ & $\always \neg \mathsf{yellow} \wedge \event (\mathsf{red} \wedge \event (\mathsf{purple} \wedge \event (\mathsf{green} \wedge \event \mathsf{red})))$ \\
 & $\varphi_{20}$ & $\event ((\mathsf{red} \vee \mathsf{green}) \wedge ((\mathsf{red} \rightarrow (\neg \mathsf{yellow} \until (\mathsf{purple} \wedge \event \mathsf{green}))) \wedge (\mathsf{green} \rightarrow (\neg \mathsf{purple} \until (\mathsf{yellow} \wedge \event \mathsf{red})))))$ \\
\midrule
\multirow{10}{*}{\rotatebox[origin=c]{90}{FrankaZoneEnv}} & $\varphi_{21}$ & $\always \neg (\mathsf{cyan} \vee \mathsf{pink}) \wedge (\event (\mathsf{red} \wedge \event (\mathsf{yellow} \wedge \event \mathsf{purple})) \vee \event (\mathsf{blue} \wedge \event (\mathsf{green} \wedge \event \mathsf{orange})))$ \\
 & $\varphi_{22}$ & $\event (\mathsf{red} \wedge \event (\mathsf{green} \wedge \event \mathsf{blue})) \wedge \event (\mathsf{orange} \wedge \event (\mathsf{pink} \wedge \event \mathsf{cyan}))$ \\
 & $\varphi_{23}$ & $\event \mathsf{yellow} \wedge \event \mathsf{purple} \wedge \event \mathsf{cyan} \wedge \event \mathsf{blue} \wedge (\neg \mathsf{cyan} \until \mathsf{yellow}) \wedge (\neg \mathsf{blue} \until \mathsf{purple})$ \\
 & $\varphi_{24}$ & $(\neg \mathsf{pink} \until \mathsf{cyan}) \wedge \event (\mathsf{pink} \wedge \event \mathsf{red})$ \\
 & $\varphi_{25}$ & $\neg (\mathsf{green} \vee \mathsf{yellow}) \until ((\mathsf{red} \vee \mathsf{purple}) \wedge (\neg (\mathsf{blue} \vee \mathsf{pink}) \until (\mathsf{cyan} \vee \mathsf{orange})))$ \\
 & $\varphi_{26}$ & $(\neg \mathsf{red} \until (\mathsf{green} \vee \mathsf{pink})) \wedge (((\mathsf{green} \rightarrow (\neg \mathsf{red} \until \mathsf{cyan})) \wedge (\mathsf{pink} \rightarrow (\neg \mathsf{red} \until \mathsf{orange}))) \until (\mathsf{red} \wedge \event \mathsf{yellow}))$ \\
 & $\varphi_{27}$ & $\always \neg (\mathsf{red} \vee \mathsf{pink}) \wedge \event ((\mathsf{green} \vee \mathsf{blue}) \wedge \event (\mathsf{yellow} \wedge \event (\mathsf{cyan} \wedge \event \mathsf{orange})))$ \\
 & $\varphi_{28}$ & $(\event \mathsf{red} \wedge \event \mathsf{green} \wedge \always \neg (\mathsf{blue} \vee \mathsf{yellow})) \vee (\event \mathsf{blue} \wedge \event \mathsf{yellow} \wedge \always \neg (\mathsf{red} \vee \mathsf{green}))$ \\
 & $\varphi_{29}$ & $\event (\mathsf{red} \wedge \event (\mathsf{green} \wedge \event (\mathsf{purple} \wedge \event (\mathsf{yellow} \wedge \event (\mathsf{blue} \wedge \event (\mathsf{orange} \wedge \event (\mathsf{cyan} \wedge \event \mathsf{pink})))))))$ \\
 & $\varphi_{30}$ & $\neg \mathsf{green} \until (\mathsf{red} \wedge (\neg \mathsf{purple} \until (\mathsf{green} \wedge (\neg \mathsf{yellow} \until (\mathsf{purple} \wedge (\neg \mathsf{blue} \until (\mathsf{yellow} \wedge (\neg \mathsf{orange} \until (\mathsf{blue} \wedge (\neg \mathsf{cyan} \until (\mathsf{orange} \wedge (\neg \mathsf{pink} \until (\mathsf{cyan} \wedge (\neg \mathsf{red} \until \mathsf{pink}))))))))))))))$ \\
\midrule
\multirow{10}{*}{\rotatebox[origin=c]{90}{Warehouse}} & $\varphi_{31}$ & $\event (\mathsf{region\_a} \wedge \event (\mathsf{crate} \wedge \event \mathsf{region\_b}))$ \\
 & $\varphi_{32}$ & $\neg (\mathsf{region\_a} \vee \mathsf{door}) \until (\mathsf{vase} \wedge \mathsf{region\_b})$ \\
 & $\varphi_{33}$ & $\neg \mathsf{door} \until (\mathsf{region\_a} \wedge (\neg \mathsf{vase} \until \mathsf{region\_b}))$ \\
 & $\varphi_{34}$ & $\neg (\mathsf{vase} \vee \mathsf{crate}) \until (\mathsf{region\_a} \wedge \event (\mathsf{door} \wedge \event \mathsf{region\_b}))$ \\
 & $\varphi_{35}$ & $\event (\mathsf{vase} \wedge \mathsf{region\_a} \wedge (\mathsf{region\_a} \until (\neg \mathsf{vase} \wedge \mathsf{region\_a})))$ \\
 & $\varphi_{36}$ & $\event (\mathsf{vase} \wedge \mathsf{crate} \wedge \mathsf{region\_b} \wedge (\mathsf{region\_b} \until (\neg \mathsf{vase} \wedge \neg \mathsf{crate} \wedge \mathsf{region\_b})))$ \\
 & $\varphi_{37}$ & $(\mathsf{crate} \rightarrow (\mathsf{crate} \until \mathsf{region\_b})) \until (\mathsf{crate} \wedge \mathsf{region\_b} \wedge (\mathsf{region\_b} \until (\neg \mathsf{crate} \wedge \mathsf{region\_b})))$ \\
 & $\varphi_{38}$ & $(\mathsf{vase} \rightarrow (\mathsf{vase} \until \mathsf{door})) \until (\event (\mathsf{crate} \wedge \mathsf{region\_b} \wedge (\mathsf{region\_b} \until (\neg \mathsf{crate} \wedge \mathsf{region\_b}))) \wedge \event (\mathsf{vase} \wedge \mathsf{door} \wedge (\mathsf{door} \until (\neg \mathsf{vase} \wedge \mathsf{door}))))$ \\
 & $\varphi_{39}$ & $(\mathsf{crate} \rightarrow (\neg (\mathsf{region\_a} \vee \mathsf{region\_b})) \until \mathsf{door}) \until (\mathsf{vase} \wedge \mathsf{door} \wedge (\mathsf{door} \until (\neg \mathsf{vase} \wedge \mathsf{door})))$ \\
 & $\varphi_{40}$ & $\event (\mathsf{crate} \wedge \mathsf{region\_a} \wedge (\mathsf{region\_a} \until (\neg \mathsf{crate} \wedge \mathsf{region\_a}))) \wedge \event (\mathsf{vase} \wedge \mathsf{door} \wedge (\mathsf{door} \until (\neg \mathsf{vase} \wedge \mathsf{door})))$ \\
\bottomrule
\toprule
\multirow{10}{*}{\rotatebox[origin=c]{90}{ZoneEnv-NM}} & $\varphi_{41}$ & $(\event \mathsf{red}) \wedge (\neg \mathsf{red} \until (\mathsf{green} \wedge \event \mathsf{yellow}))$ \\
 & $\varphi_{42}$ & $\event (\mathsf{red} \vee \mathsf{green}) \wedge \event \mathsf{yellow} \wedge \event \mathsf{purple}$ \\
 & $\varphi_{43}$ & $\neg (\mathsf{purple} \vee \mathsf{yellow}) \until (\mathsf{red} \wedge \event \mathsf{green})$ \\
 & $\varphi_{44}$ & $\neg \mathsf{green} \until ((\mathsf{red} \vee \mathsf{purple}) \wedge (\neg \mathsf{green} \until \mathsf{yellow}))$ \\
 & $\varphi_{45}$ & $((\mathsf{green} \vee \mathsf{red}) \rightarrow (\neg \mathsf{yellow} \until \mathsf{purple})) \until \mathsf{yellow}$ \\
 & $\varphi_{46}$ & $\event (\mathsf{green} \wedge (\neg \mathsf{red} \until \mathsf{yellow})) \wedge \event \mathsf{purple}$ \\
 & $\varphi_{47}$ & $\event (\mathsf{yellow} \vee \mathsf{purple} \vee \mathsf{red}) \wedge (\neg \mathsf{yellow} \until \mathsf{green})$ \\
 & $\varphi_{48}$ & $\event \mathsf{purple} \wedge \event \mathsf{green} \wedge \event \mathsf{yellow} \wedge \event \mathsf{red}$ \\
 & $\varphi_{49}$ & $\event ((\mathsf{purple} \vee \mathsf{yellow}) \wedge \event ((\mathsf{purple} \vee \mathsf{red}) \wedge \event \mathsf{green}))$ \\
 & $\varphi_{50}$ & $(\neg \mathsf{yellow} \until \mathsf{purple}) \wedge \event (\mathsf{green} \wedge \event (\mathsf{red} \wedge \event \mathsf{green}))$ \\
\bottomrule
\end{tabular}
\end{table}

\begin{table}[ht]
\caption{Infinite-horizon LTL specifications used for evaluation.}
\centering
\small
\label{tab:ltl_specs_infinite}
\begin{tabular}{c l p{12cm}}
\toprule
 & {ID} & {LTL Formula} \\
\midrule
\multirow{10}{*}{\rotatebox[origin=c]{90}{LetterWorld}} & $\psi_{1}$ & $\always \event \mathsf{a} \wedge \always \event \mathsf{b} \wedge \always \event \mathsf{c}$ \\
 & $\psi_{2}$ & $\always \event (\mathsf{a} \wedge \event \mathsf{d}) \wedge \always \event (\mathsf{b} \wedge \event \mathsf{e})$ \\
 & $\psi_{3}$ & $\always \neg \mathsf{l} \wedge \always \event \mathsf{c} \wedge \always \event \mathsf{g} \wedge \always \event \mathsf{j}$ \\
 & $\psi_{4}$ & $(\neg \mathsf{a} \until \mathsf{b}) \wedge \always \event \mathsf{a} \wedge \always \event \mathsf{f} \wedge \always \event \mathsf{g} \wedge \always \event \mathsf{h}$ \\
 & $\psi_{5}$ & $\always (\mathsf{a} \rightarrow \event \mathsf{d}) \wedge \always \event \mathsf{a} \wedge \always \event \mathsf{h}$ \\
 & $\psi_{6}$ & $\always \event (\mathsf{c} \wedge (\neg \mathsf{d} \until \mathsf{e})) \wedge \always \event \mathsf{b}$ \\
 & $\psi_{7}$ & $\event (\mathsf{a} \wedge \event (\mathsf{b} \wedge \event \mathsf{c})) \wedge \always \event \mathsf{d} \wedge \always \event \mathsf{e} \wedge \always \event \mathsf{f} \wedge \always \event \mathsf{g}$ \\
 & $\psi_{8}$ & $\always (\mathsf{b} \rightarrow (\neg \mathsf{c} \until \mathsf{d})) \wedge \always \event \mathsf{b} \wedge \always \event \mathsf{c}$ \\
 & $\psi_{9}$ & $\always \event (\mathsf{e} \wedge \event \mathsf{h}) \wedge \always \event \mathsf{k}$ \\
 & $\psi_{10}$ & $\always \event \mathsf{a} \wedge \always \event \mathsf{b} \wedge \always \event \mathsf{c} \wedge \always \event \mathsf{d} \wedge \always \neg \mathsf{l}$ \\
\midrule
\multirow{10}{*}{\rotatebox[origin=c]{90}{ZoneEnv}} & $\psi_{11}$ & $\always \event \mathsf{yellow} \wedge \always \event \mathsf{green} \wedge \always \event \mathsf{red}$ \\
 & $\psi_{12}$ & $\always \event \mathsf{red} \wedge \always \event \mathsf{green} \wedge \always \event \mathsf{yellow} \wedge \always \neg \mathsf{purple}$ \\
 & $\psi_{13}$ & $\always \event \mathsf{red} \wedge \always \event \mathsf{yellow} \wedge \always \event \mathsf{purple}$ \\
 & $\psi_{14}$ & $\always \event (\mathsf{red} \wedge \event \mathsf{green}) \wedge \always \event \mathsf{purple}$ \\
 & $\psi_{15}$ & $\always \neg \mathsf{purple} \wedge \always \event (\mathsf{red} \wedge \event \mathsf{green}) \wedge \always \event \mathsf{yellow}$ \\
 & $\psi_{16}$ & $\always (\mathsf{red} \rightarrow \event \mathsf{yellow}) \wedge \always \event \mathsf{red} \wedge \always \event \mathsf{green}$ \\
 & $\psi_{17}$ & $\always \event (\mathsf{green} \wedge (\neg \mathsf{yellow} \until \mathsf{purple})) \wedge \always \event \mathsf{red}$ \\
 & $\psi_{18}$ & $(\neg \mathsf{red} \until \mathsf{purple}) \wedge \always \event \mathsf{red} \wedge \always \event \mathsf{yellow} \wedge \always \event \mathsf{green}$ \\
 & $\psi_{19}$ & $\always (\mathsf{purple} \rightarrow (\neg \mathsf{green} \until \mathsf{yellow})) \wedge \always \event \mathsf{purple} \wedge \always \event \mathsf{green}$ \\
 & $\psi_{20}$ & $\event (\mathsf{red} \wedge \event \mathsf{green}) \wedge \always \event \mathsf{yellow} \wedge \always \event \mathsf{purple} \wedge \always \event \mathsf{red}$ \\
\midrule
\multirow{10}{*}{\rotatebox[origin=c]{90}{FrankaZoneEnv}} & $\psi_{21}$ & $\always \event \mathsf{red} \wedge \always \event \mathsf{green} \wedge \always \event \mathsf{blue} \wedge \always \event \mathsf{yellow}$ \\
 & $\psi_{22}$ & $\always \event (\mathsf{orange} \wedge \event \mathsf{pink}) \wedge \always \event (\mathsf{cyan} \wedge \event \mathsf{purple})$ \\
 & $\psi_{23}$ & $\always \neg (\mathsf{red} \vee \mathsf{pink}) \wedge \always \event \mathsf{green} \wedge \always \event \mathsf{orange} \wedge \always \event \mathsf{cyan}$ \\
 & $\psi_{24}$ & $\always \event (\mathsf{yellow} \wedge (\neg (\mathsf{blue} \vee \mathsf{red}) \until \mathsf{purple})) \wedge \always \event \mathsf{orange}$ \\
 & $\psi_{25}$ & $(\neg \mathsf{blue} \until \mathsf{cyan}) \wedge \always \event \mathsf{blue} \wedge \always \event \mathsf{pink}$ \\
 & $\psi_{26}$ & $\always (\mathsf{blue} \rightarrow \event \mathsf{yellow}) \wedge \always (\mathsf{green} \rightarrow \event \mathsf{purple}) \wedge \always \event \mathsf{blue} \wedge \always \event \mathsf{green}$ \\
 & $\psi_{27}$ & $\always \event \mathsf{cyan} \wedge \always \event \mathsf{purple} \wedge \always (\mathsf{cyan} \rightarrow (\neg \mathsf{purple} \until \mathsf{orange}))$ \\
 & $\psi_{28}$ & $\event (\mathsf{yellow} \wedge \event (\mathsf{orange} \wedge \event \mathsf{pink})) \wedge \always \event \mathsf{red} \wedge \always \event \mathsf{green}$ \\
 & $\psi_{29}$ & $\always (\mathsf{red} \rightarrow (\neg \mathsf{blue} \until \mathsf{green})) \wedge \always (\mathsf{blue} \rightarrow (\neg \mathsf{red} \until \mathsf{green})) \wedge \always \event \mathsf{red} \wedge \always \event \mathsf{blue}$ \\
 & $\psi_{30}$ & $\always \event (\mathsf{cyan} \wedge (\neg \mathsf{pink} \until \mathsf{purple})) \wedge \always \event (\mathsf{pink} \wedge (\neg \mathsf{cyan} \until \mathsf{yellow}))$ \\
\midrule
\multirow{10}{*}{\rotatebox[origin=c]{90}{Warehouse}} & $\psi_{31}$ & $\always \event \mathsf{region\_a} \wedge \always \event \mathsf{region\_b} \wedge \always \event \mathsf{door}$ \\
 & $\psi_{32}$ & $\always \event (\mathsf{vase} \wedge \event \neg \mathsf{vase}) \wedge \always \event (\mathsf{crate} \wedge \event \neg \mathsf{crate}) \wedge \always \event \mathsf{door} \wedge \always \neg \mathsf{region\_a}$ \\
 & $\psi_{33}$ & $\event (\mathsf{vase} \wedge \mathsf{region\_a} \wedge (\mathsf{region\_a} \until (\neg \mathsf{vase} \wedge \mathsf{region\_a}))) \wedge \always \event \mathsf{door} \wedge \always \event \mathsf{region\_b}$ \\
 & $\psi_{34}$ & $\always \event (\mathsf{vase} \wedge \mathsf{region\_a} \wedge (\mathsf{region\_a} \until (\neg \mathsf{vase} \wedge \mathsf{region\_a}))) \wedge \always \event (\mathsf{crate} \wedge \mathsf{region\_b} \wedge (\mathsf{region\_b} \until (\neg \mathsf{crate} \wedge \mathsf{region\_b})))$ \\
 & $\psi_{35}$ & $\always \event \mathsf{region\_a} \wedge \always \event \mathsf{region\_b} \wedge \always \event \mathsf{door} \wedge \always (\mathsf{door} \rightarrow \event \mathsf{region\_a})$ \\
 & $\psi_{36}$ & $\always \event (\mathsf{vase} \wedge \event (\neg \mathsf{vase} \wedge \mathsf{region\_a})) \wedge \always \event \mathsf{region\_b}$ \\
 & $\psi_{37}$ & $\always \event (\mathsf{crate} \wedge \event (\neg \mathsf{crate} \wedge \mathsf{region\_b})) \wedge \always \event \mathsf{door}$ \\
 & $\psi_{38}$ & $\always \neg \mathsf{region\_b} \wedge \always \event \mathsf{region\_a} \wedge \always \event \mathsf{door} \wedge \always \event (\mathsf{vase} \wedge \event \neg \mathsf{vase})$ \\
 & $\psi_{39}$ & $\always \event (\mathsf{region\_a} \wedge \event \mathsf{region\_b}) \wedge \always \event \mathsf{door}$ \\
 & $\psi_{40}$ & $\always (\mathsf{vase} \rightarrow \event (\neg \mathsf{vase} \wedge \mathsf{region\_a})) \wedge \always \event \mathsf{vase} \wedge \always \event \mathsf{door}$ \\
\bottomrule
\end{tabular}
\end{table}

\begin{table}[ht]
\caption{Reach-stay LTL specifications used for evaluation.}
\centering
\small
\label{tab:ltl_specs_reach_stay}
\begin{tabular}{c l p{12cm}}
\toprule
 & {ID} & {LTL Formula} \\
\midrule
\multirow{10}{*}{\rotatebox[origin=c]{90}{ZoneEnv}} & $\chi_{1}$ & $\event \always \mathsf{red}$ \\
 & $\chi_{2}$ & $\event \always \mathsf{red} \wedge \event (\mathsf{yellow} \wedge \event \mathsf{green})$ \\
 & $\chi_{3}$ & $\event \always \mathsf{purple} \wedge \always \neg \mathsf{yellow}$ \\
 & $\chi_{4}$ & $\always ((\mathsf{green} \vee \mathsf{yellow}) \rightarrow \event \mathsf{red}) \wedge \event \always (\mathsf{green} \vee \mathsf{purple})$ \\
 & $\chi_{5}$ & $\event \always \mathsf{green}$ \\
 & $\chi_{6}$ & $\event \always \mathsf{yellow}$ \\
 & $\chi_{7}$ & $\event \always \mathsf{green} \wedge \always \neg \mathsf{red}$ \\
 & $\chi_{8}$ & $\event \always \mathsf{purple} \wedge \event (\mathsf{green} \wedge \event \mathsf{red})$ \\
 & $\chi_{9}$ & $\event \always \mathsf{purple} \wedge \always \neg \mathsf{green}$ \\
 & $\chi_{10}$ & $\always (\mathsf{yellow} \rightarrow \event \mathsf{red}) \wedge \event \always \mathsf{purple}$ \\
\midrule
\multirow{10}{*}{\rotatebox[origin=c]{90}{Warehouse}} & $\chi_{11}$ & $\event \always \mathsf{region\_a}$ \\
 & $\chi_{12}$ & $\event (\mathsf{vase} \wedge \mathsf{region\_b} \wedge (\mathsf{region\_b} \until (\neg \mathsf{vase} \wedge \mathsf{region\_b}))) \wedge \event \always \mathsf{region\_b}$ \\
 & $\chi_{13}$ & $\event \always \mathsf{region\_b}$ \\
 & $\chi_{14}$ & $\event \always \mathsf{vase} \wedge \always \neg \mathsf{region\_a}$ \\
 & $\chi_{15}$ & $\event \mathsf{vase} \wedge \event \always \mathsf{region\_a}$ \\
 & $\chi_{16}$ & $\event \mathsf{crate} \wedge \event \always \mathsf{region\_b}$ \\
 & $\chi_{17}$ & $\event \always \mathsf{region\_a} \wedge \always \neg \mathsf{crate}$ \\
 & $\chi_{18}$ & $\event \mathsf{vase} \wedge \event \always \mathsf{region\_b}$ \\
 & $\chi_{19}$ & $\event (\mathsf{crate} \wedge \mathsf{region\_a} \wedge (\mathsf{region\_a} \until (\neg \mathsf{crate} \wedge \mathsf{region\_a}))) \wedge \event \always \mathsf{region\_a}$ \\
 & $\chi_{20}$ & $\event \always \mathsf{region\_b} \wedge \event (\mathsf{crate} \wedge \event \neg \mathsf{crate})$ \\
\bottomrule
\end{tabular}
\end{table}

\begingroup
\newcommand{\appendixscore}[2]{#1\ensuremath{_{\pm #2}}}
\newcommand{\appendixbest}[2]{\appendixscore{\textbf{#1}}{#2}}
\newcommand{\appendixsecond}[2]{\appendixscore{\underline{#1}}{#2}}
\begin{table*}[t]
\centering
\caption{Per-specification finite-horizon benchmark results (success rate), averaged 
over 512 evaluation episodes, with $95\%$ Student-$t$ CIs over training seeds. 
Best results are \textbf{bold}, second best \underline{underlined}, --- denotes unsupported configurations.}
\label{tab:appendix-results-finite}
\scriptsize
\setlength{\tabcolsep}{2pt}
\resizebox{.9\textwidth}{!}{%
\begin{tabular}{@{}c l *{7}{c}@{}}
\toprule
 & $\varphi$ & LTL2Action & GCRL-LTL & DeepLTL & GenZ-LTL$^\dagger$ & \makecell{GenZ-LTL\\(unreduced obs.)} & SemLTL & StructLTL \\
\midrule
\multirow{10}{*}{\rotatebox[origin=c]{90}{LetterWorld}} & $\varphi_{1}$ & \appendixscore{0.62}{0.14} & \appendixscore{0.99}{0.00} & \appendixsecond{1.00}{0.00} & \appendixscore{0.99}{0.00} & \appendixscore{1.00}{0.00} & \appendixbest{1.00}{0.00} & \appendixscore{1.00}{0.00} \\
 & $\varphi_{2}$ & \appendixscore{0.74}{0.15} & \appendixsecond{0.99}{0.00} & \appendixscore{0.96}{0.01} & \appendixbest{1.00}{0.00} & \appendixscore{0.98}{0.00} & \appendixscore{0.95}{0.01} & \appendixscore{0.97}{0.01} \\
 & $\varphi_{3}$ & \appendixscore{0.45}{0.13} & \appendixbest{1.00}{0.00} & \appendixscore{1.00}{0.00} & \appendixbest{1.00}{0.00} & \appendixscore{1.00}{0.00} & \appendixsecond{1.00}{0.00} & \appendixbest{1.00}{0.00} \\
 & $\varphi_{4}$ & \appendixscore{0.52}{0.20} & \appendixsecond{0.98}{0.01} & \appendixscore{0.95}{0.01} & \appendixbest{0.99}{0.00} & \appendixscore{0.97}{0.01} & \appendixscore{0.94}{0.01} & \appendixscore{0.95}{0.01} \\
 & $\varphi_{5}$ & \appendixscore{0.02}{0.01} & \appendixsecond{0.92}{0.01} & \appendixscore{0.73}{0.03} & \appendixbest{1.00}{0.00} & \appendixscore{0.85}{0.02} & \appendixscore{0.57}{0.07} & \appendixscore{0.79}{0.04} \\
 & $\varphi_{6}$ & \appendixscore{0.00}{0.00} & \appendixscore{0.72}{0.01} & \appendixscore{0.53}{0.07} & \appendixbest{0.97}{0.01} & \appendixsecond{0.77}{0.02} & \appendixscore{0.32}{0.04} & \appendixscore{0.62}{0.03} \\
 & $\varphi_{7}$ & \appendixscore{0.63}{0.15} & \appendixscore{1.00}{0.00} & \appendixsecond{1.00}{0.00} & \appendixscore{0.99}{0.00} & \appendixscore{1.00}{0.00} & \appendixbest{1.00}{0.00} & \appendixscore{1.00}{0.00} \\
 & $\varphi_{8}$ & \appendixscore{0.90}{0.03} & \appendixbest{1.00}{0.00} & \appendixscore{1.00}{0.00} & \appendixbest{1.00}{0.00} & \appendixscore{0.99}{0.00} & \appendixscore{0.81}{0.06} & \appendixsecond{1.00}{0.00} \\
 & $\varphi_{9}$ & \appendixscore{0.01}{0.01} & \appendixsecond{0.97}{0.01} & \appendixscore{0.79}{0.02} & \appendixbest{1.00}{0.00} & \appendixscore{0.89}{0.01} & \appendixscore{0.62}{0.07} & \appendixscore{0.83}{0.03} \\
 & $\varphi_{10}$ & \appendixscore{0.27}{0.10} & \appendixscore{0.85}{0.02} & \appendixscore{0.78}{0.02} & \appendixbest{1.00}{0.00} & \appendixsecond{0.89}{0.02} & \appendixscore{0.70}{0.02} & \appendixscore{0.81}{0.02} \\
\midrule
\multirow{10}{*}{\rotatebox[origin=c]{90}{ZoneEnv}} & $\varphi_{11}$ & \appendixscore{0.33}{0.11} & \appendixscore{0.99}{0.00} & \appendixscore{0.96}{0.02} & \appendixbest{1.00}{0.00} & \appendixsecond{1.00}{0.00} & \appendixscore{0.97}{0.02} & \appendixscore{0.98}{0.02} \\
 & $\varphi_{12}$ & \appendixscore{0.24}{0.21} & \appendixscore{0.95}{0.01} & \appendixscore{0.94}{0.01} & \appendixbest{1.00}{0.00} & \appendixscore{0.97}{0.01} & \appendixscore{0.92}{0.01} & \appendixsecond{0.97}{0.01} \\
 & $\varphi_{13}$ & \appendixscore{0.68}{0.13} & \appendixsecond{1.00}{0.00} & \appendixscore{0.98}{0.01} & \appendixbest{1.00}{0.00} & \appendixsecond{1.00}{0.00} & \appendixscore{0.98}{0.01} & \appendixscore{0.99}{0.00} \\
 & $\varphi_{14}$ & \appendixscore{0.68}{0.15} & \appendixscore{0.92}{0.01} & \appendixscore{0.91}{0.02} & \appendixbest{0.99}{0.00} & \appendixscore{0.93}{0.01} & \appendixscore{0.91}{0.02} & \appendixsecond{0.94}{0.01} \\
 & $\varphi_{15}$ & \appendixscore{0.79}{0.16} & \appendixscore{0.93}{0.01} & \appendixscore{0.94}{0.01} & \appendixbest{1.00}{0.00} & \appendixscore{0.95}{0.01} & \appendixscore{0.90}{0.05} & \appendixsecond{0.97}{0.00} \\
 & $\varphi_{16}$ & \appendixscore{0.90}{0.04} & \appendixscore{0.99}{0.00} & \appendixscore{0.96}{0.01} & \appendixbest{1.00}{0.00} & \appendixscore{0.99}{0.01} & \appendixscore{0.98}{0.01} & \appendixsecond{0.99}{0.01} \\
 & $\varphi_{17}$ & \appendixscore{0.06}{0.06} & \appendixscore{0.88}{0.02} & \appendixscore{0.83}{0.07} & \appendixbest{0.99}{0.00} & \appendixscore{0.93}{0.02} & \appendixscore{0.83}{0.05} & \appendixsecond{0.94}{0.02} \\
 & $\varphi_{18}$ & \appendixscore{0.47}{0.19} & \appendixscore{0.85}{0.01} & \appendixscore{0.81}{0.03} & \appendixbest{0.98}{0.00} & \appendixscore{0.88}{0.02} & \appendixscore{0.83}{0.01} & \appendixsecond{0.88}{0.01} \\
 & $\varphi_{19}$ & \appendixscore{0.03}{0.06} & \appendixscore{0.83}{0.02} & \appendixscore{0.84}{0.07} & \appendixbest{0.99}{0.00} & \appendixsecond{0.92}{0.02} & \appendixscore{0.71}{0.09} & \appendixscore{0.88}{0.08} \\
 & $\varphi_{20}$ & \appendixscore{0.17}{0.07} & \appendixscore{1.00}{0.00} & \appendixscore{0.98}{0.01} & \appendixbest{1.00}{0.00} & \appendixsecond{1.00}{0.00} & \appendixscore{0.32}{0.07} & \appendixscore{0.98}{0.02} \\
\midrule
\multirow{10}{*}{\rotatebox[origin=c]{90}{FrankaZoneEnv}} & $\varphi_{21}$ & \appendixscore{0.02}{0.01} & \appendixscore{0.50}{0.26} & \appendixscore{0.39}{0.18} & \appendixbest{1.00}{0.00} & \appendixscore{0.45}{0.21} & \appendixscore{0.50}{0.13} & \appendixsecond{0.88}{0.01} \\
 & $\varphi_{22}$ & \appendixscore{0.00}{0.00} & \appendixscore{0.56}{0.31} & \appendixscore{0.76}{0.14} & \appendixbest{1.00}{0.00} & \appendixscore{0.38}{0.26} & \appendixscore{0.48}{0.08} & \appendixsecond{0.99}{0.00} \\
 & $\varphi_{23}$ & \appendixscore{0.02}{0.01} & \appendixscore{0.54}{0.28} & \appendixscore{0.54}{0.18} & \appendixbest{1.00}{0.00} & \appendixscore{0.44}{0.23} & \appendixscore{0.62}{0.10} & \appendixsecond{0.94}{0.01} \\
 & $\varphi_{24}$ & \appendixscore{0.07}{0.03} & \appendixscore{0.66}{0.29} & \appendixscore{0.62}{0.20} & \appendixbest{1.00}{0.00} & \appendixscore{0.55}{0.27} & \appendixscore{0.89}{0.04} & \appendixsecond{0.97}{0.01} \\
 & $\varphi_{25}$ & \appendixscore{0.14}{0.05} & \appendixscore{0.72}{0.19} & \appendixscore{0.50}{0.10} & \appendixbest{1.00}{0.00} & \appendixscore{0.71}{0.20} & \appendixscore{0.83}{0.02} & \appendixsecond{0.97}{0.01} \\
 & $\varphi_{26}$ & \appendixscore{0.01}{0.01} & \appendixscore{0.53}{0.30} & \appendixscore{0.50}{0.21} & \appendixbest{1.00}{0.00} & \appendixscore{0.45}{0.26} & \appendixscore{0.28}{0.05} & \appendixsecond{0.95}{0.01} \\
 & $\varphi_{27}$ & \appendixscore{0.00}{0.00} & \appendixscore{0.43}{0.27} & \appendixscore{0.25}{0.16} & \appendixbest{1.00}{0.00} & \appendixscore{0.32}{0.18} & \appendixscore{0.44}{0.15} & \appendixsecond{0.81}{0.01} \\
 & $\varphi_{28}$ & \appendixscore{0.13}{0.03} & \appendixscore{0.70}{0.20} & \appendixscore{0.63}{0.14} & \appendixbest{1.00}{0.00} & \appendixscore{0.60}{0.19} & \appendixscore{0.52}{0.19} & \appendixsecond{0.95}{0.01} \\
 & $\varphi_{29}$ & \appendixscore{0.00}{0.00} & \appendixscore{0.47}{0.33} & \appendixscore{0.58}{0.21} & \appendixbest{1.00}{0.00} & \appendixscore{0.31}{0.25} & \appendixscore{0.34}{0.11} & \appendixsecond{0.99}{0.00} \\
 & $\varphi_{30}$ & \appendixscore{0.00}{0.00} & \appendixscore{0.39}{0.28} & \appendixscore{0.35}{0.26} & \appendixbest{1.00}{0.00} & \appendixscore{0.27}{0.23} & \appendixscore{0.23}{0.08} & \appendixsecond{0.95}{0.01} \\
\midrule
\multirow{10}{*}{\rotatebox[origin=c]{90}{Warehouse}} & $\varphi_{31}$ & \appendixscore{0.97}{0.02} & --- & \appendixsecond{0.98}{0.01} & --- & \appendixscore{0.96}{0.03} & \appendixscore{0.95}{0.10} & \appendixbest{0.99}{0.01} \\
 & $\varphi_{32}$ & \appendixscore{0.41}{0.26} & --- & \appendixsecond{0.94}{0.02} & --- & \appendixscore{0.91}{0.03} & \appendixscore{0.92}{0.03} & \appendixbest{0.95}{0.01} \\
 & $\varphi_{33}$ & \appendixscore{0.19}{0.22} & --- & \appendixsecond{0.97}{0.03} & --- & \appendixscore{0.95}{0.04} & \appendixbest{0.99}{0.01} & \appendixscore{0.94}{0.08} \\
 & $\varphi_{34}$ & \appendixscore{0.21}{0.10} & --- & \appendixbest{0.93}{0.02} & --- & \appendixscore{0.74}{0.27} & \appendixscore{0.78}{0.08} & \appendixsecond{0.79}{0.14} \\
 & $\varphi_{35}$ & \appendixscore{0.62}{0.33} & --- & \appendixscore{0.76}{0.22} & --- & \appendixsecond{0.99}{0.01} & \appendixscore{0.98}{0.02} & \appendixbest{0.99}{0.00} \\
 & $\varphi_{36}$ & \appendixscore{0.52}{0.31} & --- & \appendixscore{0.72}{0.20} & --- & \appendixsecond{0.96}{0.01} & \appendixscore{0.83}{0.16} & \appendixbest{0.99}{0.01} \\
 & $\varphi_{37}$ & \appendixscore{0.12}{0.10} & --- & \appendixscore{0.85}{0.11} & --- & \appendixsecond{0.97}{0.02} & \appendixscore{0.75}{0.11} & \appendixbest{0.98}{0.01} \\
 & $\varphi_{38}$ & \appendixscore{0.11}{0.19} & --- & \appendixscore{0.60}{0.24} & --- & \appendixsecond{0.63}{0.31} & \appendixscore{0.41}{0.14} & \appendixbest{0.88}{0.22} \\
 & $\varphi_{39}$ & \appendixscore{0.11}{0.14} & --- & \appendixscore{0.77}{0.20} & --- & \appendixscore{0.66}{0.33} & \appendixsecond{0.80}{0.12} & \appendixbest{0.95}{0.03} \\
 & $\varphi_{40}$ & \appendixscore{0.11}{0.20} & --- & \appendixscore{0.51}{0.27} & --- & \appendixscore{0.64}{0.31} & \appendixbest{0.94}{0.04} & \appendixsecond{0.88}{0.22} \\
\bottomrule
\end{tabular}%
}
\end{table*}

\begin{table*}[t]
\centering
\caption{Per-specification infinite-horizon benchmark results (completed accepting cycles), averaged 
over 512 evaluation episodes, with $95\%$ Student-$t$ CIs over training seeds. 
Best results are \textbf{bold}, second best \underline{underlined}, --- denotes unsupported configurations.}
\label{tab:appendix-results-infinite}
\scriptsize
\setlength{\tabcolsep}{2pt}
\resizebox{.9\textwidth}{!}{%
\begin{tabular}{@{}c l *{6}{c}@{}}
\toprule
 & $\psi$ & GCRL-LTL & DeepLTL & GenZ-LTL$^\dagger$ & \makecell{GenZ-LTL\\(unreduced obs.)} & SemLTL & StructLTL \\
\midrule
\multirow{10}{*}{\rotatebox[origin=c]{90}{LetterWorld}} & $\psi_{1}$ & \appendixsecond{9.4}{0.1} & \appendixscore{6.8}{0.3} & \appendixbest{9.8}{0.1} & \appendixscore{6.9}{0.1} & \appendixscore{6.0}{0.2} & \appendixscore{6.7}{0.4} \\
 & $\psi_{2}$ & \appendixsecond{6.4}{0.0} & \appendixscore{4.3}{0.2} & \appendixbest{6.7}{0.1} & \appendixscore{4.5}{0.0} & \appendixscore{4.1}{0.1} & \appendixscore{4.3}{0.2} \\
 & $\psi_{3}$ & \appendixsecond{8.8}{0.5} & \appendixscore{5.3}{0.4} & \appendixbest{9.8}{0.1} & \appendixscore{6.2}{0.1} & \appendixscore{4.5}{0.3} & \appendixscore{5.7}{0.3} \\
 & $\psi_{4}$ & \appendixsecond{6.3}{0.0} & \appendixscore{4.0}{0.3} & \appendixbest{6.7}{0.1} & \appendixscore{4.5}{0.0} & \appendixscore{3.3}{0.2} & \appendixscore{4.1}{0.2} \\
 & $\psi_{5}$ & \appendixbest{9.4}{0.3} & \appendixscore{6.3}{0.9} & \appendixscore{6.4}{0.3} & \appendixscore{5.2}{0.3} & \appendixscore{6.0}{0.3} & \appendixsecond{7.1}{0.8} \\
 & $\psi_{6}$ & \appendixsecond{9.3}{0.1} & \appendixscore{6.4}{0.5} & \appendixbest{9.9}{0.1} & \appendixscore{6.8}{0.2} & \appendixscore{4.2}{0.6} & \appendixscore{6.8}{0.4} \\
 & $\psi_{7}$ & \appendixsecond{5.6}{0.1} & \appendixscore{3.6}{0.2} & \appendixbest{5.9}{0.0} & \appendixscore{3.8}{0.1} & \appendixscore{2.7}{0.2} & \appendixscore{3.5}{0.2} \\
 & $\psi_{8}$ & \appendixsecond{9.7}{0.1} & \appendixscore{6.2}{0.4} & \appendixbest{10.5}{0.1} & \appendixscore{7.0}{0.1} & \appendixscore{2.1}{0.3} & \appendixscore{6.8}{0.2} \\
 & $\psi_{9}$ & \appendixsecond{9.4}{0.1} & \appendixscore{6.6}{0.4} & \appendixbest{9.9}{0.1} & \appendixscore{7.0}{0.2} & \appendixscore{6.2}{0.4} & \appendixscore{6.8}{0.3} \\
 & $\psi_{10}$ & \appendixsecond{6.0}{0.1} & \appendixscore{3.4}{0.2} & \appendixbest{6.7}{0.1} & \appendixscore{3.9}{0.1} & \appendixscore{2.7}{0.2} & \appendixscore{3.5}{0.2} \\
\midrule
\multirow{10}{*}{\rotatebox[origin=c]{90}{ZoneEnv}} & $\psi_{11}$ & \appendixsecond{6.0}{0.1} & \appendixscore{3.3}{0.4} & \appendixbest{6.7}{0.3} & \appendixscore{5.1}{0.2} & \appendixscore{3.3}{0.2} & \appendixscore{3.9}{0.4} \\
 & $\psi_{12}$ & \appendixscore{4.5}{0.2} & \appendixscore{2.3}{0.4} & \appendixbest{6.6}{0.3} & \appendixsecond{4.7}{0.1} & \appendixscore{2.8}{0.1} & \appendixscore{3.0}{0.3} \\
 & $\psi_{13}$ & \appendixsecond{6.0}{0.1} & \appendixscore{3.2}{0.3} & \appendixbest{6.7}{0.3} & \appendixscore{5.1}{0.2} & \appendixscore{3.2}{0.1} & \appendixscore{3.9}{0.5} \\
 & $\psi_{14}$ & \appendixsecond{5.9}{0.1} & \appendixscore{3.3}{0.5} & \appendixbest{6.6}{0.2} & \appendixscore{5.1}{0.2} & \appendixscore{3.1}{0.1} & \appendixscore{3.6}{0.6} \\
 & $\psi_{15}$ & \appendixscore{4.4}{0.2} & \appendixscore{2.2}{0.4} & \appendixbest{6.6}{0.3} & \appendixsecond{4.6}{0.2} & \appendixscore{2.8}{0.2} & \appendixscore{2.9}{0.3} \\
 & $\psi_{16}$ & \appendixbest{6.9}{0.1} & \appendixscore{3.5}{0.6} & \appendixsecond{5.1}{0.9} & \appendixscore{4.7}{0.7} & \appendixscore{3.2}{0.2} & \appendixscore{4.2}{0.5} \\
 & $\psi_{17}$ & \appendixsecond{5.7}{0.1} & \appendixscore{3.1}{0.5} & \appendixbest{6.6}{0.3} & \appendixscore{5.0}{0.2} & \appendixscore{2.3}{0.3} & \appendixscore{3.7}{0.5} \\
 & $\psi_{18}$ & \appendixsecond{5.6}{0.1} & \appendixscore{3.0}{0.3} & \appendixbest{6.7}{0.3} & \appendixscore{5.0}{0.1} & \appendixscore{2.5}{0.2} & \appendixscore{3.7}{0.4} \\
 & $\psi_{19}$ & \appendixsecond{6.3}{0.1} & \appendixscore{3.3}{0.4} & \appendixbest{7.3}{0.2} & \appendixscore{5.5}{0.2} & \appendixscore{1.2}{0.4} & \appendixscore{4.3}{0.2} \\
 & $\psi_{20}$ & \appendixsecond{5.7}{0.1} & \appendixscore{2.9}{0.3} & \appendixbest{6.4}{0.3} & \appendixscore{4.8}{0.1} & \appendixscore{2.1}{0.2} & \appendixscore{3.5}{0.4} \\
\midrule
\multirow{10}{*}{\rotatebox[origin=c]{90}{FrankaZoneEnv}} & $\psi_{21}$ & \appendixscore{5.3}{4.0} & \appendixscore{4.0}{2.7} & \appendixbest{24.6}{0.3} & \appendixscore{3.4}{2.9} & \appendixsecond{15.1}{1.7} & \appendixscore{11.9}{0.3} \\
 & $\psi_{22}$ & \appendixscore{5.4}{4.0} & \appendixscore{3.9}{2.6} & \appendixbest{23.9}{0.3} & \appendixscore{3.6}{3.0} & \appendixsecond{14.3}{1.2} & \appendixscore{11.5}{0.4} \\
 & $\psi_{23}$ & \appendixscore{5.7}{4.6} & \appendixscore{2.2}{2.5} & \appendixbest{32.1}{0.6} & \appendixscore{3.3}{3.0} & \appendixsecond{12.7}{3.5} & \appendixscore{10.7}{0.9} \\
 & $\psi_{24}$ & \appendixscore{7.3}{5.5} & \appendixscore{5.0}{3.6} & \appendixbest{33.9}{0.5} & \appendixscore{4.4}{3.3} & \appendixscore{8.2}{2.0} & \appendixsecond{15.3}{0.4} \\
 & $\psi_{25}$ & \appendixscore{17.1}{12.7} & \appendixscore{8.6}{5.8} & \appendixbest{67.7}{2.3} & \appendixscore{9.9}{7.9} & \appendixsecond{35.7}{2.9} & \appendixscore{27.8}{1.1} \\
 & $\psi_{26}$ & \appendixscore{5.6}{3.4} & \appendixscore{4.1}{1.7} & \appendixsecond{13.5}{2.1} & \appendixscore{3.0}{1.4} & \appendixbest{15.8}{1.4} & \appendixscore{11.1}{0.5} \\
 & $\psi_{27}$ & \appendixscore{7.7}{5.1} & \appendixscore{5.1}{3.3} & \appendixbest{33.7}{0.8} & \appendixscore{5.2}{3.5} & \appendixscore{2.8}{1.6} & \appendixsecond{14.0}{0.8} \\
 & $\psi_{28}$ & \appendixscore{16.3}{12.4} & \appendixscore{8.2}{6.3} & \appendixbest{68.1}{1.9} & \appendixscore{9.5}{7.6} & \appendixsecond{28.9}{4.7} & \appendixscore{28.0}{1.2} \\
 & $\psi_{29}$ & \appendixscore{6.2}{4.0} & \appendixscore{3.7}{2.5} & \appendixbest{28.4}{0.7} & \appendixscore{4.2}{3.0} & \appendixscore{1.1}{1.1} & \appendixsecond{11.2}{0.5} \\
 & $\psi_{30}$ & \appendixscore{4.8}{3.5} & \appendixscore{3.5}{2.6} & \appendixbest{23.6}{0.3} & \appendixscore{4.9}{3.4} & \appendixscore{1.3}{0.4} & \appendixsecond{11.3}{0.2} \\
\midrule
\multirow{10}{*}{\rotatebox[origin=c]{90}{Warehouse}} & $\psi_{31}$ & --- & \appendixsecond{2.3}{0.4} & --- & \appendixscore{1.6}{0.6} & \appendixscore{0.2}{0.2} & \appendixbest{2.5}{0.2} \\
 & $\psi_{32}$ & --- & \appendixbest{4.0}{3.3} & --- & \appendixsecond{2.6}{1.5} & \appendixscore{0.3}{0.4} & \appendixscore{2.1}{1.2} \\
 & $\psi_{33}$ & --- & \appendixsecond{1.9}{0.7} & --- & \appendixscore{1.9}{0.9} & \appendixscore{0.3}{0.2} & \appendixbest{3.0}{0.1} \\
 & $\psi_{34}$ & --- & \appendixscore{1.7}{0.9} & --- & \appendixsecond{3.0}{0.3} & \appendixscore{0.1}{0.1} & \appendixbest{3.8}{0.2} \\
 & $\psi_{35}$ & --- & \appendixsecond{2.3}{0.3} & --- & \appendixscore{1.5}{0.7} & \appendixscore{0.2}{0.2} & \appendixbest{2.8}{0.1} \\
 & $\psi_{36}$ & --- & \appendixsecond{3.7}{0.5} & --- & \appendixscore{3.2}{0.9} & \appendixscore{0.2}{0.2} & \appendixbest{4.4}{0.3} \\
 & $\psi_{37}$ & --- & \appendixsecond{2.5}{0.5} & --- & \appendixscore{1.9}{0.8} & \appendixscore{0.0}{0.0} & \appendixbest{2.9}{0.1} \\
 & $\psi_{38}$ & --- & \appendixsecond{1.9}{0.6} & --- & \appendixscore{1.8}{0.7} & \appendixscore{0.3}{0.2} & \appendixbest{2.6}{0.5} \\
 & $\psi_{39}$ & --- & \appendixsecond{2.2}{0.4} & --- & \appendixscore{1.6}{0.6} & \appendixscore{0.2}{0.2} & \appendixbest{2.4}{0.3} \\
 & $\psi_{40}$ & --- & \appendixsecond{3.7}{0.6} & --- & \appendixscore{3.0}{0.9} & \appendixscore{0.1}{0.1} & \appendixbest{4.2}{0.2} \\
\bottomrule
\end{tabular}%
}
\end{table*}

\begin{table*}[t]
\centering
\caption{Per-specification reach-stay benchmark results (completed accepting cycles), averaged 
over 512 evaluation episodes, with $95\%$ Student-$t$ CIs over training seeds. 
Best results are \textbf{bold}, second best \underline{underlined}, --- denotes unsupported configurations.}
\label{tab:appendix-results-reach-stay}
\scriptsize
\setlength{\tabcolsep}{2pt}
\resizebox{.9\textwidth}{!}{%
\begin{tabular}{@{}c l *{6}{c}@{}}
\toprule
 & $\chi$ & GCRL-LTL & DeepLTL & GenZ-LTL$^\dagger$ & \makecell{GenZ-LTL\\(unreduced obs.)} & SemLTL & StructLTL \\
\midrule
\multirow{10}{*}{\rotatebox[origin=c]{90}{ZoneEnv}} & $\chi_{1}$ & \appendixscore{24.2}{5.0} & \appendixbest{645.1}{85.6} & \appendixscore{548.0}{57.2} & \appendixscore{340.6}{119.3} & \appendixsecond{615.7}{38.5} & \appendixscore{608.4}{61.8} \\
 & $\chi_{2}$ & \appendixscore{51.2}{16.2} & \appendixsecond{497.2}{61.5} & \appendixscore{471.7}{61.7} & \appendixscore{307.4}{99.8} & \appendixscore{445.5}{22.1} & \appendixbest{501.2}{56.8} \\
 & $\chi_{3}$ & \appendixscore{21.6}{3.3} & \appendixscore{512.8}{79.5} & \appendixscore{533.1}{51.9} & \appendixscore{317.6}{103.0} & \appendixbest{591.4}{67.9} & \appendixsecond{575.4}{63.8} \\
 & $\chi_{4}$ & \appendixscore{28.4}{6.7} & \appendixsecond{520.9}{85.7} & \appendixscore{495.6}{37.5} & \appendixscore{303.1}{104.9} & \appendixscore{285.3}{101.4} & \appendixbest{571.0}{64.9} \\
 & $\chi_{5}$ & \appendixscore{25.9}{4.3} & \appendixsecond{635.4}{51.3} & \appendixscore{541.4}{51.2} & \appendixscore{333.8}{129.0} & \appendixbest{679.1}{64.9} & \appendixscore{593.8}{100.8} \\
 & $\chi_{6}$ & \appendixscore{18.9}{2.6} & \appendixscore{513.4}{113.3} & \appendixscore{299.4}{94.5} & \appendixscore{341.8}{104.1} & \appendixbest{639.0}{66.7} & \appendixsecond{626.8}{97.9} \\
 & $\chi_{7}$ & \appendixscore{22.8}{4.3} & \appendixsecond{603.1}{44.7} & \appendixscore{511.4}{52.9} & \appendixscore{321.6}{127.6} & \appendixbest{631.5}{71.5} & \appendixscore{573.9}{106.0} \\
 & $\chi_{8}$ & \appendixscore{60.3}{12.9} & \appendixscore{438.9}{71.3} & \appendixscore{477.3}{60.3} & \appendixscore{306.9}{93.7} & \appendixsecond{495.3}{53.7} & \appendixbest{511.6}{61.2} \\
 & $\chi_{9}$ & \appendixscore{22.1}{4.1} & \appendixscore{513.6}{88.2} & \appendixscore{511.2}{54.7} & \appendixscore{318.4}{104.1} & \appendixbest{605.0}{61.7} & \appendixsecond{577.8}{76.8} \\
 & $\chi_{10}$ & \appendixscore{26.1}{3.6} & \appendixscore{521.1}{81.9} & \appendixsecond{530.7}{53.6} & \appendixscore{325.0}{103.9} & \appendixscore{341.4}{90.3} & \appendixbest{586.7}{59.3} \\
\midrule
\multirow{10}{*}{\rotatebox[origin=c]{90}{Warehouse}} & $\chi_{11}$ & --- & \appendixsecond{860.6}{63.4} & --- & \appendixscore{128.6}{48.6} & \appendixscore{823.1}{24.9} & \appendixbest{872.2}{35.1} \\
 & $\chi_{12}$ & --- & \appendixscore{337.4}{216.5} & --- & \appendixscore{66.4}{27.7} & \appendixsecond{538.6}{86.2} & \appendixbest{800.2}{64.3} \\
 & $\chi_{13}$ & --- & \appendixscore{600.5}{224.8} & --- & \appendixscore{70.9}{32.3} & \appendixsecond{759.4}{78.6} & \appendixbest{867.8}{48.0} \\
 & $\chi_{14}$ & --- & \appendixsecond{656.5}{175.3} & --- & \appendixscore{193.9}{79.6} & \appendixscore{622.1}{157.2} & \appendixbest{712.9}{86.0} \\
 & $\chi_{15}$ & --- & \appendixsecond{791.5}{65.1} & --- & \appendixscore{117.9}{45.3} & \appendixscore{745.9}{45.2} & \appendixbest{815.9}{36.6} \\
 & $\chi_{16}$ & --- & \appendixscore{563.4}{197.4} & --- & \appendixscore{75.6}{30.2} & \appendixsecond{712.9}{40.1} & \appendixbest{808.9}{41.8} \\
 & $\chi_{17}$ & --- & \appendixbest{819.9}{83.0} & --- & \appendixscore{115.1}{34.7} & \appendixscore{613.4}{120.6} & \appendixsecond{788.7}{142.4} \\
 & $\chi_{18}$ & --- & \appendixscore{568.4}{210.6} & --- & \appendixscore{69.1}{26.2} & \appendixsecond{671.6}{76.1} & \appendixbest{797.3}{65.4} \\
 & $\chi_{19}$ & --- & \appendixscore{414.2}{222.1} & --- & \appendixscore{121.2}{58.1} & \appendixsecond{592.0}{85.8} & \appendixbest{821.8}{30.0} \\
 & $\chi_{20}$ & --- & \appendixscore{568.4}{202.8} & --- & \appendixscore{67.2}{23.8} & \appendixsecond{655.7}{61.6} & \appendixbest{813.0}{42.8} \\
\bottomrule
\end{tabular}%
}
\end{table*}
\endgroup

\begin{table*}[t]
\centering
\caption{Training and model hyperparameters for LetterWorld. Values spanning multiple columns are shared by those methods; N/A marks settings that do not apply to a method.}
\label{tab:hyperparameters_letter_world}
\resizebox{\textwidth}{!}{%
\begin{tabular}{llcccccc}
\toprule
Category & Hyperparameter & LTL2Action & SemLTL & DeepLTL & StructLTL & GCRL-LTL & GenZ-LTL \\
\midrule
\multirow{13}{*}{PPO}
& Total environment steps & \multicolumn{6}{c}{\spanval{$2 \times 10^{7}$}} \\
& Environments & \multicolumn{6}{c}{\spanval{16}} \\
& Steps/update & \multicolumn{6}{c}{\spanval{128}} \\
& Minibatches & \multicolumn{6}{c}{\spanval{8}} \\
& Update epochs & \multicolumn{6}{c}{\spanval{8}} \\
& Discount ($\gamma$) & \multicolumn{6}{c}{\spanval{0.94}} \\
& GAE lambda ($\lambda$) & \multicolumn{6}{c}{\spanval{0.95}} \\
& Clip epsilon & \multicolumn{6}{c}{\spanval{0.2}} \\
& Entropy coef. & 0.01 & \multicolumn{5}{c}{\spanval{0.05}} \\
& Value func.\ coef. & \multicolumn{6}{c}{\spanval{0.5}} \\
& Learning rate & \multicolumn{6}{c}{\spanval{$3 \times 10^{-4}$}} \\
& Max grad norm & \multicolumn{6}{c}{\spanval{0.5}} \\
& Adam epsilon & \multicolumn{6}{c}{\spanval{$10^{-8}$}} \\
\midrule
\multirow{7}{*}{Safe PPO}
& Cost discount ($\gamma_c$) & \multicolumn{5}{c}{\spanval{N/A}} & 0.94 \\
& Cost value func.\ coef. & \multicolumn{5}{c}{\spanval{N/A}} & 1.0 \\
& Lagrangian coef. & \multicolumn{5}{c}{\spanval{N/A}} & 1.0 \\
& Target cost & \multicolumn{5}{c}{\spanval{N/A}} & $-0.1$ \\
& Min Lagrangian & \multicolumn{5}{c}{\spanval{N/A}} & $10^{-3}$ \\
& Max Lagrangian & \multicolumn{5}{c}{\spanval{N/A}} & 3.0 \\
& Target KL & \multicolumn{5}{c}{\spanval{N/A}} & 0.015 \\
\midrule
\multirow{4}{*}{Curriculum}
& Samples per stage & \multicolumn{5}{c}{\spanval{$10^{4}$}} & $10^{5}$ \\
& Episode window & N/A & \multicolumn{3}{c}{\spanval{256}} & \multicolumn{2}{c}{\spanval{N/A}} \\
& Adoption prob. & N/A & \multicolumn{3}{c}{\spanval{0.1}} & \multicolumn{2}{c}{\spanval{N/A}} \\
& Min coverage & N/A & \multicolumn{3}{c}{\spanval{0.9}} & \multicolumn{2}{c}{\spanval{N/A}} \\
\midrule
\multirow{3}{*}{Env Net}
& Channels & \multicolumn{6}{c}{\spanval{[16, 32, 64]}} \\
& Kernel size & \multicolumn{6}{c}{\spanval{[2, 2]}} \\
& Activation & \multicolumn{6}{c}{\spanval{ReLU}} \\
\midrule
\multirow{2}{*}{Actor}
& Hidden sizes & \multicolumn{6}{c}{\spanval{[64, 64, 64]}} \\
& Activation & \multicolumn{6}{c}{\spanval{ReLU}} \\
\midrule
\multirow{2}{*}{Critic}
& Hidden sizes & \multicolumn{6}{c}{\spanval{[64, 64]}} \\
& Activation & \multicolumn{6}{c}{\spanval{Tanh}} \\
\midrule
\multirow{2}{*}{Cost Critic}
& Hidden sizes & \multicolumn{5}{c}{\spanval{N/A}} & [64, 64] \\
& Activation & \multicolumn{5}{c}{\spanval{N/A}} & Tanh \\
\midrule
\multirow{2}{*}{Lagrangian Net}
& Hidden sizes & \multicolumn{5}{c}{\spanval{N/A}} & [64, 64] \\
& Activation & \multicolumn{5}{c}{\spanval{N/A}} & Softplus \\
\midrule
\multirow{12}{*}{LTL Encoder}
& Embedding dim. & 32 & 64 & \multicolumn{3}{c}{\spanval{32}} & N/A \\
& RGCN layers & 8 & \multicolumn{5}{c}{\spanval{N/A}} \\
& RGCN activation & Tanh & \multicolumn{5}{c}{\spanval{N/A}} \\
& Semantic input size & N/A & 762 & \multicolumn{4}{c}{\spanval{N/A}} \\
& Sequence model & \multicolumn{2}{c}{\spanval{N/A}} & GRU & ALiBi attention & \multicolumn{2}{c}{\spanval{N/A}} \\
& Deep sets hidden sizes & \multicolumn{2}{c}{\spanval{N/A}} & [32, 32] & \multicolumn{3}{c}{\spanval{N/A}} \\
& Deep sets output size & \multicolumn{2}{c}{\spanval{N/A}} & 32 & \multicolumn{3}{c}{\spanval{N/A}} \\
& Deep sets activation & \multicolumn{2}{c}{\spanval{N/A}} & ReLU & \multicolumn{3}{c}{\spanval{N/A}} \\
& Clause net output size & \multicolumn{3}{c}{\spanval{N/A}} & 32 & \multicolumn{2}{c}{\spanval{N/A}} \\
& Disjunct net output size & \multicolumn{3}{c}{\spanval{N/A}} & 32 & \multicolumn{2}{c}{\spanval{N/A}} \\
& Attention hidden dim. & \multicolumn{3}{c}{\spanval{N/A}} & 32 & \multicolumn{2}{c}{\spanval{N/A}} \\
& ALiBi slope & \multicolumn{3}{c}{\spanval{N/A}} & 0.5 & \multicolumn{2}{c}{\spanval{N/A}} \\
\midrule
\multirow{6}{*}{GCVF}
& Samples & \multicolumn{4}{c}{\spanval{N/A}} & $2 \times 10^{5}$ & N/A \\
& Batch size & \multicolumn{4}{c}{\spanval{N/A}} & 512 & N/A \\
& Learning rate & \multicolumn{4}{c}{\spanval{N/A}} & $4 \times 10^{-4}$ & N/A \\
& Epochs & \multicolumn{4}{c}{\spanval{N/A}} & 100 & N/A \\
& Environments & \multicolumn{4}{c}{\spanval{N/A}} & 16 & N/A \\
& Steps/env & \multicolumn{4}{c}{\spanval{N/A}} & 2,048 & N/A \\
\midrule
\multirow{2}{*}{Environment}
& Precomputed resets & \multicolumn{6}{c}{\spanval{10,240}} \\
& Max episode length & \multicolumn{6}{c}{\spanval{75}} \\
\bottomrule
\end{tabular}
}
\end{table*}

\begin{table*}[t]
\centering
\caption{Training and model hyperparameters for ZoneEnv and ZoneEnv-NM. Values spanning multiple columns are shared by those methods; N/A marks settings that do not apply to a method.}
\label{tab:hyperparameters_zone_env}
\resizebox{\textwidth}{!}{%
\begin{tabular}{llcccccc}
\toprule
Category & Hyperparameter & LTL2Action & SemLTL & DeepLTL & StructLTL & GCRL-LTL & GenZ-LTL \\
\midrule
\multirow{13}{*}{PPO}
& Total environment steps & \multicolumn{6}{c}{\spanval{$2 \times 10^{7}$}} \\
& Environments & \multicolumn{6}{c}{\spanval{16}} \\
& Steps/update & \multicolumn{6}{c}{\spanval{4,096}} \\
& Minibatches & \multicolumn{6}{c}{\spanval{32}} \\
& Update epochs & \multicolumn{6}{c}{\spanval{10}} \\
& Discount ($\gamma$) & \multicolumn{6}{c}{\spanval{0.998}} \\
& GAE lambda ($\lambda$) & \multicolumn{6}{c}{\spanval{0.95}} \\
& Clip epsilon & \multicolumn{6}{c}{\spanval{0.2}} \\
& Entropy coef. & \multicolumn{6}{c}{\spanval{0.003}} \\
& Value func.\ coef. & \multicolumn{5}{c}{\spanval{0.5}} & 1.0 \\
& Learning rate & \multicolumn{6}{c}{\spanval{$3 \times 10^{-4}$}} \\
& Max grad norm & \multicolumn{6}{c}{\spanval{0.5}} \\
& Adam epsilon & \multicolumn{6}{c}{\spanval{$10^{-8}$}} \\
\midrule
\multirow{7}{*}{Safe PPO}
& Cost discount ($\gamma_c$) & \multicolumn{5}{c}{\spanval{N/A}} & 0.998 \\
& Cost value func.\ coef. & \multicolumn{5}{c}{\spanval{N/A}} & 1.0 \\
& Lagrangian coef. & \multicolumn{5}{c}{\spanval{N/A}} & 1.0 \\
& Target cost & \multicolumn{5}{c}{\spanval{N/A}} & $-0.05$ \\
& Min Lagrangian & \multicolumn{5}{c}{\spanval{N/A}} & 0.01 \\
& Max Lagrangian & \multicolumn{5}{c}{\spanval{N/A}} & 5.0 \\
& Target KL & \multicolumn{5}{c}{\spanval{N/A}} & 0.015 \\
\midrule
\multirow{4}{*}{Curriculum}
& Samples per stage & \multicolumn{5}{c}{\spanval{$10^{4}$}} & $10^{5}$ \\
& Episode window & \multicolumn{4}{c}{\spanval{256}} & \multicolumn{2}{c}{\spanval{N/A}} \\
& Adoption prob. & \multicolumn{4}{c}{\spanval{0.1}} & \multicolumn{2}{c}{\spanval{N/A}} \\
& Min coverage & \multicolumn{4}{c}{\spanval{0.9}} & \multicolumn{2}{c}{\spanval{N/A}} \\
\midrule
\multirow{3}{*}{Env Net}
& Hidden sizes & \multicolumn{6}{c}{\spanval{[128]}} \\
& Output size & \multicolumn{6}{c}{\spanval{64}} \\
& Activation & \multicolumn{6}{c}{\spanval{Tanh}} \\
\midrule
\multirow{3}{*}{Actor}
& Hidden sizes & \multicolumn{6}{c}{\spanval{[64, 64, 64]}} \\
& Activation & \multicolumn{6}{c}{\spanval{ReLU}} \\
& Output activation & \multicolumn{4}{c}{\spanval{Tanh}} & N/A & Tanh \\
\midrule
\multirow{2}{*}{Critic}
& Hidden sizes & \multicolumn{6}{c}{\spanval{[64, 64]}} \\
& Activation & \multicolumn{5}{c}{\spanval{Tanh}} & Softplus \\
\midrule
\multirow{2}{*}{Cost Critic}
& Hidden sizes & \multicolumn{5}{c}{\spanval{N/A}} & [64, 64] \\
& Activation & \multicolumn{5}{c}{\spanval{N/A}} & Tanh \\
\midrule
\multirow{2}{*}{Lagrangian Net}
& Hidden sizes & \multicolumn{5}{c}{\spanval{N/A}} & [64, 64] \\
& Activation & \multicolumn{5}{c}{\spanval{N/A}} & Softplus \\
\midrule
\multirow{12}{*}{LTL Encoder}
& Embedding dim. & \multicolumn{2}{c}{\spanval{32}} & \multicolumn{3}{c}{\spanval{16}} & N/A \\
& RGCN layers & 8 & \multicolumn{5}{c}{\spanval{N/A}} \\
& RGCN activation & Tanh & \multicolumn{5}{c}{\spanval{N/A}} \\
& Semantic input size & N/A & 138 & \multicolumn{4}{c}{\spanval{N/A}} \\
& Sequence model & \multicolumn{2}{c}{\spanval{N/A}} & GRU & ALiBi attention & \multicolumn{2}{c}{\spanval{N/A}} \\
& Deep sets hidden sizes & \multicolumn{2}{c}{\spanval{N/A}} & [32] & \multicolumn{3}{c}{\spanval{N/A}} \\
& Deep sets output size & \multicolumn{2}{c}{\spanval{N/A}} & 16 & \multicolumn{3}{c}{\spanval{N/A}} \\
& Deep sets activation & \multicolumn{2}{c}{\spanval{N/A}} & ReLU & \multicolumn{3}{c}{\spanval{N/A}} \\
& Clause net output size & \multicolumn{3}{c}{\spanval{N/A}} & 16 & \multicolumn{2}{c}{\spanval{N/A}} \\
& Disjunct net output size & \multicolumn{3}{c}{\spanval{N/A}} & 16 & \multicolumn{2}{c}{\spanval{N/A}} \\
& Attention hidden dim. & \multicolumn{3}{c}{\spanval{N/A}} & 32 & \multicolumn{2}{c}{\spanval{N/A}} \\
& ALiBi slope & \multicolumn{3}{c}{\spanval{N/A}} & 0.5 & \multicolumn{2}{c}{\spanval{N/A}} \\
\midrule
\multirow{6}{*}{GCVF}
& Samples & \multicolumn{4}{c}{\spanval{N/A}} & $2 \times 10^{5}$ & N/A \\
& Batch size & \multicolumn{4}{c}{\spanval{N/A}} & 512 & N/A \\
& Learning rate & \multicolumn{4}{c}{\spanval{N/A}} & $4 \times 10^{-4}$ & N/A \\
& Epochs & \multicolumn{4}{c}{\spanval{N/A}} & 100 & N/A \\
& Environments & \multicolumn{4}{c}{\spanval{N/A}} & 16 & N/A \\
& Steps/env & \multicolumn{4}{c}{\spanval{N/A}} & 2,048 & N/A \\
\midrule
\multirow{2}{*}{Environment}
& Precomputed resets & \multicolumn{6}{c}{\spanval{10,240}} \\
& Max episode length & \multicolumn{6}{c}{\spanval{1,000}} \\
\bottomrule
\end{tabular}
}
\end{table*}

\begin{table*}[t]
\centering
\caption{Training and model hyperparameters for FrankaZoneEnv. Values spanning multiple columns are shared by those methods; N/A marks settings that do not apply to a method. The table covers FrankaZoneEnv-$k$ for $k \in \{8, 9, 10, 11, 12\}$. Hyperparameters that vary with $k$ are given as functions of $k$.}
\label{tab:hyperparameters_franka_zone_env}
\resizebox{\textwidth}{!}{%
\begin{tabular}{llcccccc}
\toprule
Category & Hyperparameter & LTL2Action & SemLTL & DeepLTL & StructLTL & GCRL-LTL & GenZ-LTL \\
\midrule
\multirow{13}{*}{PPO}
& Total environment steps & \multicolumn{6}{c}{\spanval{$2 \times 10^{8}$}} \\
& Environments & \multicolumn{6}{c}{\spanval{2,048}} \\
& Steps/update & \multicolumn{6}{c}{\spanval{64}} \\
& Minibatches & \multicolumn{6}{c}{\spanval{4}} \\
& Update epochs & \multicolumn{6}{c}{\spanval{5}} \\
& Discount ($\gamma$) & \multicolumn{6}{c}{\spanval{0.995}} \\
& GAE lambda ($\lambda$) & \multicolumn{6}{c}{\spanval{0.95}} \\
& Clip epsilon & \multicolumn{6}{c}{\spanval{0.2}} \\
& Entropy coef. & \multicolumn{6}{c}{\spanval{0.003}} \\
& Value func.\ coef. & \multicolumn{5}{c}{\spanval{0.5}} & 1.0 \\
& Learning rate & \multicolumn{6}{c}{\spanval{$5 \times 10^{-4}$}} \\
& Max grad norm & \multicolumn{6}{c}{\spanval{1.0}} \\
& Adam epsilon & \multicolumn{6}{c}{\spanval{$10^{-8}$}} \\
\midrule
\multirow{7}{*}{Safe PPO}
& Cost discount ($\gamma_c$) & \multicolumn{5}{c}{\spanval{N/A}} & 0.995 \\
& Cost value func.\ coef. & \multicolumn{5}{c}{\spanval{N/A}} & 1.0 \\
& Lagrangian coef. & \multicolumn{5}{c}{\spanval{N/A}} & 1.0 \\
& Target cost & \multicolumn{5}{c}{\spanval{N/A}} & $-0.05$ \\
& Min Lagrangian & \multicolumn{5}{c}{\spanval{N/A}} & 0.01 \\
& Max Lagrangian & \multicolumn{5}{c}{\spanval{N/A}} & 5.0 \\
& Target KL & \multicolumn{5}{c}{\spanval{N/A}} & 0.015 \\
\midrule
\multirow{4}{*}{Curriculum}
& Samples per stage & \multicolumn{6}{c}{\spanval{$10^{4}$}} \\
& Episode window & \multicolumn{4}{c}{\spanval{512}} & \multicolumn{2}{c}{\spanval{N/A}} \\
& Adoption prob. & \multicolumn{4}{c}{\spanval{0.1}} & \multicolumn{2}{c}{\spanval{N/A}} \\
& Min coverage & \multicolumn{4}{c}{\spanval{0.9}} & \multicolumn{2}{c}{\spanval{N/A}} \\
\midrule
\multirow{3}{*}{Env Net}
& Hidden sizes & \multicolumn{6}{c}{\spanval{[128]}} \\
& Output size & \multicolumn{6}{c}{\spanval{64}} \\
& Activation & \multicolumn{6}{c}{\spanval{Tanh}} \\
\midrule
\multirow{3}{*}{Actor}
& Hidden sizes & \multicolumn{6}{c}{\spanval{[128, 128]}} \\
& Activation & \multicolumn{6}{c}{\spanval{ReLU}} \\
& Output activation & \multicolumn{4}{c}{\spanval{Tanh}} & N/A & Tanh \\
\midrule
\multirow{2}{*}{Critic}
& Hidden sizes & \multicolumn{6}{c}{\spanval{[128, 128]}} \\
& Activation & \multicolumn{5}{c}{\spanval{Tanh}} & Softplus \\
\midrule
\multirow{2}{*}{Cost Critic}
& Hidden sizes & \multicolumn{5}{c}{\spanval{N/A}} & [128, 128] \\
& Activation & \multicolumn{5}{c}{\spanval{N/A}} & Tanh \\
\midrule
\multirow{2}{*}{Lagrangian Net}
& Hidden sizes & \multicolumn{5}{c}{\spanval{N/A}} & [128, 128] \\
& Activation & \multicolumn{5}{c}{\spanval{N/A}} & Softplus \\
\midrule
\multirow{12}{*}{LTL Encoder}
& Embedding dim. & \multicolumn{2}{c}{\spanval{32}} & \multicolumn{3}{c}{\spanval{16}} & N/A \\
& RGCN layers & 8 & \multicolumn{5}{c}{\spanval{N/A}} \\
& RGCN activation & Tanh & \multicolumn{5}{c}{\spanval{N/A}} \\
& Semantic input size & N/A & $4k^{2} + 14k + 18$ & \multicolumn{4}{c}{\spanval{N/A}} \\
& Sequence model & \multicolumn{2}{c}{\spanval{N/A}} & GRU & ALiBi attention & \multicolumn{2}{c}{\spanval{N/A}} \\
& Deep sets hidden sizes & \multicolumn{2}{c}{\spanval{N/A}} & [32] & \multicolumn{3}{c}{\spanval{N/A}} \\
& Deep sets output size & \multicolumn{2}{c}{\spanval{N/A}} & 16 & \multicolumn{3}{c}{\spanval{N/A}} \\
& Deep sets activation & \multicolumn{2}{c}{\spanval{N/A}} & ReLU & \multicolumn{3}{c}{\spanval{N/A}} \\
& Clause net output size & \multicolumn{3}{c}{\spanval{N/A}} & 16 & \multicolumn{2}{c}{\spanval{N/A}} \\
& Disjunct net output size & \multicolumn{3}{c}{\spanval{N/A}} & 16 & \multicolumn{2}{c}{\spanval{N/A}} \\
& Attention hidden dim. & \multicolumn{3}{c}{\spanval{N/A}} & 32 & \multicolumn{2}{c}{\spanval{N/A}} \\
& ALiBi slope & \multicolumn{3}{c}{\spanval{N/A}} & 0.5 & \multicolumn{2}{c}{\spanval{N/A}} \\
\midrule
\multirow{6}{*}{GCVF}
& Samples & \multicolumn{4}{c}{\spanval{N/A}} & $2 \times 10^{5}$ & N/A \\
& Batch size & \multicolumn{4}{c}{\spanval{N/A}} & 512 & N/A \\
& Learning rate & \multicolumn{4}{c}{\spanval{N/A}} & $4 \times 10^{-4}$ & N/A \\
& Epochs & \multicolumn{4}{c}{\spanval{N/A}} & 100 & N/A \\
& Environments & \multicolumn{4}{c}{\spanval{N/A}} & 16 & N/A \\
& Steps/env & \multicolumn{4}{c}{\spanval{N/A}} & 2,048 & N/A \\
\midrule
\multirow{3}{*}{Environment}
& Precomputed resets & \multicolumn{6}{c}{\spanval{10,240}} \\
& Max episode length (train) & \multicolumn{6}{c}{\spanval{200}} \\
& Max episode length (eval) & \multicolumn{6}{c}{\spanval{1,000}} \\
\bottomrule
\end{tabular}
}
\end{table*}

\begin{table*}[t]
\centering
\caption{Training and model hyperparameters for Warehouse. Values spanning multiple columns are shared by those methods; N/A marks settings that do not apply to a method.}
\label{tab:hyperparameters_warehouse}
\resizebox{\textwidth}{!}{%
\begin{tabular}{llccccc}
\toprule
Category & Hyperparameter & LTL2Action & SemLTL & DeepLTL & StructLTL & GenZ-LTL \\
\midrule
\multirow{13}{*}{PPO}
& Total environment steps & \multicolumn{5}{c}{\spanval{$2 \times 10^{8}$}} \\
& Environments & \multicolumn{5}{c}{\spanval{1,024}} \\
& Steps/update & \multicolumn{5}{c}{\spanval{128}} \\
& Minibatches & \multicolumn{5}{c}{\spanval{4}} \\
& Update epochs & \multicolumn{5}{c}{\spanval{5}} \\
& Discount ($\gamma$) & \multicolumn{5}{c}{\spanval{0.998}} \\
& GAE lambda ($\lambda$) & \multicolumn{5}{c}{\spanval{0.95}} \\
& Clip epsilon & \multicolumn{5}{c}{\spanval{0.2}} \\
& Entropy coef. & \multicolumn{5}{c}{\spanval{0.003}} \\
& Value func.\ coef. & \multicolumn{4}{c}{\spanval{0.5}} & 1.0 \\
& Learning rate & \multicolumn{5}{c}{\spanval{$5 \times 10^{-4}$}} \\
& Max grad norm & \multicolumn{5}{c}{\spanval{1.0}} \\
& Adam epsilon & \multicolumn{5}{c}{\spanval{$10^{-8}$}} \\
\midrule
\multirow{7}{*}{Safe PPO}
& Cost discount ($\gamma_c$) & \multicolumn{4}{c}{\spanval{N/A}} & 0.998 \\
& Cost value func.\ coef. & \multicolumn{4}{c}{\spanval{N/A}} & 1.0 \\
& Lagrangian coef. & \multicolumn{4}{c}{\spanval{N/A}} & 1.0 \\
& Target cost & \multicolumn{4}{c}{\spanval{N/A}} & $-0.1$ \\
& Min Lagrangian & \multicolumn{4}{c}{\spanval{N/A}} & $10^{-3}$ \\
& Max Lagrangian & \multicolumn{4}{c}{\spanval{N/A}} & 3.0 \\
& Target KL & \multicolumn{4}{c}{\spanval{N/A}} & 0.015 \\
\midrule
\multirow{4}{*}{Curriculum}
& Samples per stage & \multicolumn{4}{c}{\spanval{$10^{4}$}} & $10^{5}$ \\
& Episode window & \multicolumn{4}{c}{\spanval{256}} & N/A \\
& Adoption prob. & \multicolumn{4}{c}{\spanval{0.1}} & N/A \\
& Min coverage & \multicolumn{4}{c}{\spanval{0.9}} & N/A \\
\midrule
\multirow{3}{*}{Env Net}
& Hidden sizes & \multicolumn{5}{c}{\spanval{[128]}} \\
& Output size & \multicolumn{5}{c}{\spanval{64}} \\
& Activation & \multicolumn{5}{c}{\spanval{Tanh}} \\
\midrule
\multirow{3}{*}{Actor}
& Hidden sizes & \multicolumn{5}{c}{\spanval{[64, 64, 64]}} \\
& Activation & \multicolumn{5}{c}{\spanval{ReLU}} \\
& Output activation & \multicolumn{5}{c}{\spanval{Tanh}} \\
\midrule
\multirow{2}{*}{Critic}
& Hidden sizes & \multicolumn{5}{c}{\spanval{[64, 64]}} \\
& Activation & \multicolumn{4}{c}{\spanval{Tanh}} & Softplus \\
\midrule
\multirow{2}{*}{Cost Critic}
& Hidden sizes & \multicolumn{4}{c}{\spanval{N/A}} & [64, 64] \\
& Activation & \multicolumn{4}{c}{\spanval{N/A}} & Tanh \\
\midrule
\multirow{2}{*}{Lagrangian Net}
& Hidden sizes & \multicolumn{4}{c}{\spanval{N/A}} & [64, 64] \\
& Activation & \multicolumn{4}{c}{\spanval{N/A}} & Softplus \\
\midrule
\multirow{12}{*}{LTL Encoder}
& Embedding dim. & 16 & 32 & \multicolumn{2}{c}{\spanval{16}} & N/A \\
& RGCN layers & 8 & \multicolumn{4}{c}{\spanval{N/A}} \\
& RGCN activation & Tanh & \multicolumn{4}{c}{\spanval{N/A}} \\
& Semantic input size & N/A & 488 & \multicolumn{3}{c}{\spanval{N/A}} \\
& Sequence model & \multicolumn{2}{c}{\spanval{N/A}} & GRU & ALiBi attention & N/A \\
& Deep sets hidden sizes & \multicolumn{2}{c}{\spanval{N/A}} & [32] & \multicolumn{2}{c}{\spanval{N/A}} \\
& Deep sets output size & \multicolumn{2}{c}{\spanval{N/A}} & 16 & \multicolumn{2}{c}{\spanval{N/A}} \\
& Deep sets activation & \multicolumn{2}{c}{\spanval{N/A}} & ReLU & \multicolumn{2}{c}{\spanval{N/A}} \\
& Clause net output size & \multicolumn{3}{c}{\spanval{N/A}} & 16 & N/A \\
& Disjunct net output size & \multicolumn{3}{c}{\spanval{N/A}} & 16 & N/A \\
& Attention hidden dim. & \multicolumn{3}{c}{\spanval{N/A}} & 32 & N/A \\
& ALiBi slope & \multicolumn{3}{c}{\spanval{N/A}} & 5.0 & N/A \\
\midrule
\multirow{2}{*}{Environment}
& Precomputed resets & \multicolumn{5}{c}{\spanval{10,240}} \\
& Max episode length & \multicolumn{5}{c}{\spanval{1,000}} \\
\bottomrule
\end{tabular}
}
\end{table*}

\begin{table*}[t]
\centering
\caption{Training and model hyperparameters for ConveyorWorld and ConveyorWorldSimple. Values spanning multiple columns are shared by those methods; N/A marks settings that do not apply to a method. The table covers ConveyorWorld-$k$ for $k \in \{1, 2, 3, 4, 5, 6, 7, 8, 9, 10\}$ and ConveyorWorldSimple-$k$ for $k \in \{1, 2, 4, 8, 16, 32\}$. Hyperparameters that vary with $k$ are given as functions of $k$. Note that environment resets are not precomputed for these environments (since they are trivially cheap), and the training ``curriculum'' only consists of a single stage (i.e.\ no curriculum) for all methods.}
\label{tab:hyperparameters_conveyor}
\resizebox{\textwidth}{!}{%
\begin{tabular}{llcccccc}
\toprule
Category & Hyperparameter & LTL2Action & SemLTL & DeepLTL & StructLTL & GCRL-LTL & GenZ-LTL \\
\midrule
\multirow{13}{*}{PPO}
& Total environment steps & \multicolumn{6}{c}{\spanval{$2 \times 10^{6}$}} \\
& Environments & \multicolumn{6}{c}{\spanval{256}} \\
& Steps/update & \multicolumn{6}{c}{\spanval{64}} \\
& Minibatches & \multicolumn{6}{c}{\spanval{4}} \\
& Update epochs & \multicolumn{6}{c}{\spanval{8}} \\
& Discount ($\gamma$) & \multicolumn{6}{c}{\spanval{0.9}} \\
& GAE lambda ($\lambda$) & \multicolumn{6}{c}{\spanval{0.95}} \\
& Clip epsilon & \multicolumn{6}{c}{\spanval{0.2}} \\
& Entropy coef. & \multicolumn{6}{c}{\spanval{0.05}} \\
& Value func.\ coef. & \multicolumn{6}{c}{\spanval{0.5}} \\
& Learning rate & \multicolumn{6}{c}{\spanval{$3 \times 10^{-4}$}} \\
& Max grad norm & \multicolumn{6}{c}{\spanval{0.5}} \\
& Adam epsilon & \multicolumn{6}{c}{\spanval{$10^{-8}$}} \\
\midrule
\multirow{7}{*}{Safe PPO}
& Cost discount ($\gamma_c$) & \multicolumn{5}{c}{\spanval{N/A}} & 0.9 \\
& Cost value func.\ coef. & \multicolumn{5}{c}{\spanval{N/A}} & 1.0 \\
& Lagrangian coef. & \multicolumn{5}{c}{\spanval{N/A}} & 1.0 \\
& Target cost & \multicolumn{5}{c}{\spanval{N/A}} & $-0.1$ \\
& Min Lagrangian & \multicolumn{5}{c}{\spanval{N/A}} & $10^{-3}$ \\
& Max Lagrangian & \multicolumn{5}{c}{\spanval{N/A}} & 3.0 \\
& Target KL & \multicolumn{5}{c}{\spanval{N/A}} & 0.015 \\
\midrule
Curriculum
& Samples per stage & \multicolumn{6}{c}{\spanval{$10^{4}$}} \\
\midrule
\multirow{3}{*}{Env Net}
& Hidden sizes & \multicolumn{5}{c}{\spanval{[]}} & [128] \\
& Output size & \multicolumn{5}{c}{\spanval{4}} & 64 \\
& Activation & \multicolumn{6}{c}{\spanval{Tanh}} \\
\midrule
\multirow{2}{*}{Actor}
& Hidden sizes & \multicolumn{6}{c}{\spanval{[64]}} \\
& Activation & \multicolumn{6}{c}{\spanval{ReLU}} \\
\midrule
\multirow{2}{*}{Critic}
& Hidden sizes & \multicolumn{6}{c}{\spanval{[64]}} \\
& Activation & \multicolumn{5}{c}{\spanval{Tanh}} & Softplus \\
\midrule
\multirow{2}{*}{Cost Critic}
& Hidden sizes & \multicolumn{5}{c}{\spanval{N/A}} & [64] \\
& Activation & \multicolumn{5}{c}{\spanval{N/A}} & Tanh \\
\midrule
\multirow{2}{*}{Lagrangian Net}
& Hidden sizes & \multicolumn{5}{c}{\spanval{N/A}} & [64] \\
& Activation & \multicolumn{5}{c}{\spanval{N/A}} & Softplus \\
\midrule
\multirow{12}{*}{LTL Encoder}
& Embedding dim. & 32 & \multicolumn{4}{c}{\spanval{16}} & N/A \\
& RGCN layers & $2k + 1$ & \multicolumn{5}{c}{\spanval{N/A}} \\
& RGCN activation & Tanh & \multicolumn{5}{c}{\spanval{N/A}} \\
& Semantic input size & N/A & $4k^{2} + 22k + 36$ & \multicolumn{4}{c}{\spanval{N/A}} \\
& Sequence model & \multicolumn{2}{c}{\spanval{N/A}} & GRU & ALiBi attention & \multicolumn{2}{c}{\spanval{N/A}} \\
& Deep sets hidden sizes & \multicolumn{2}{c}{\spanval{N/A}} & [] & \multicolumn{3}{c}{\spanval{N/A}} \\
& Deep sets output size & \multicolumn{2}{c}{\spanval{N/A}} & 32 & \multicolumn{3}{c}{\spanval{N/A}} \\
& Deep sets activation & \multicolumn{2}{c}{\spanval{N/A}} & ReLU & \multicolumn{3}{c}{\spanval{N/A}} \\
& Clause net output size & \multicolumn{3}{c}{\spanval{N/A}} & 16 & \multicolumn{2}{c}{\spanval{N/A}} \\
& Disjunct net output size & \multicolumn{3}{c}{\spanval{N/A}} & 16 & \multicolumn{2}{c}{\spanval{N/A}} \\
& Attention hidden dim. & \multicolumn{3}{c}{\spanval{N/A}} & 32 & \multicolumn{2}{c}{\spanval{N/A}} \\
& ALiBi slope & \multicolumn{3}{c}{\spanval{N/A}} & 0.5 & \multicolumn{2}{c}{\spanval{N/A}} \\
\midrule
\multirow{6}{*}{GCVF}
& Samples & \multicolumn{4}{c}{\spanval{N/A}} & $2 \times 10^{5}$ & N/A \\
& Batch size & \multicolumn{4}{c}{\spanval{N/A}} & 512 & N/A \\
& Learning rate & \multicolumn{4}{c}{\spanval{N/A}} & $4 \times 10^{-4}$ & N/A \\
& Epochs & \multicolumn{4}{c}{\spanval{N/A}} & 100 & N/A \\
& Environments & \multicolumn{4}{c}{\spanval{N/A}} & 16 & N/A \\
& Steps/env & \multicolumn{4}{c}{\spanval{N/A}} & 2,048 & N/A \\
\midrule
Environment
& Max episode length & \multicolumn{6}{c}{\spanval{50 (ConveyorWorld), $\max(50, 4k)$ (ConveyorWorldSimple)}} \\
\bottomrule
\end{tabular}
}
\end{table*}

\end{document}